\documentclass[11pt, a4paper, twocolumn, copyright, gdm]{google}

\usepackage[authoryear, sort&compress, round]{natbib}
\usepackage{enumitem}    
\usepackage{xcolor}      
\usepackage{tabularx}
\usepackage{booktabs}
\usepackage{xltabular}
\usepackage{longtable}
\usepackage{float}
\usepackage{graphicx}
\usepackage{url}
\usepackage{listingsutf8}
\usepackage[utf8]{inputenc}

\DeclareUnicodeCharacter{2192}{\ensuremath{\rightarrow}}
\DeclareUnicodeCharacter{2190}{\ensuremath{\leftarrow}}
\DeclareUnicodeCharacter{21D2}{\ensuremath{\Rightarrow}}


\uselogo{} 

\title{Tailored to you: longitudinal effects of personalising language models}

\correspondingauthor{tailored-paper@google.com}

\author[*,1]{Canfer Akbulut}
\author[*,1]{Justine Breuch}
\author[*,1]{Arianna Manzini}
\author[1]{Lujain Ibrahim}
\author[1]{Matija Franklin}
\author[1]{Roma Patel}
\author[1]{Iason Gabriel}
\author[1]{Kristian Lum}
\author[1]{Laura Weidinger}

\affil[*]{Equal contributions}
\affil[1]{Google DeepMind}

\begin{abstract}
Interest in developing \textit{personalised} language models is rapidly growing. While personalisation is often viewed as a mechanism to better serve diverse user needs, the effects of sustained interactions with personalised models on people's perception of and behaviour toward AI remain poorly understood. Most critically, downstream consequences outside the immediate human--AI interaction loop, such as effects on users' self-perceptions and interpersonal relationships, remain largely unexamined. In this study, we recruited 992 participants to complete daily advice-seeking interactions with language models over the course of five days, comparing outcomes from a non-personalised baseline against two personalisation approaches: \textit{memory-based} (conditioned on prior conversational history) and \textit{survey-based} (conditioned on information collected through a pre-study intake survey). We find that several changes in human-AI interaction over time are driven primarily by \textit{repeated exposure} rather than personalisation itself. However, participants interacting with personalised models experienced differences in advice-seeking and information-sharing attitudes and behaviours. We conclude by highlighting the nuanced effects of different personalisation approaches on interaction outcomes, and discussing the implications of these findings for the responsible design and deployment of personalised AI systems.
\end{abstract}

\begin{document}

\maketitle

\section{Introduction}


\begin{figure*}[t]
    \centering
    \includegraphics[width=\textwidth]{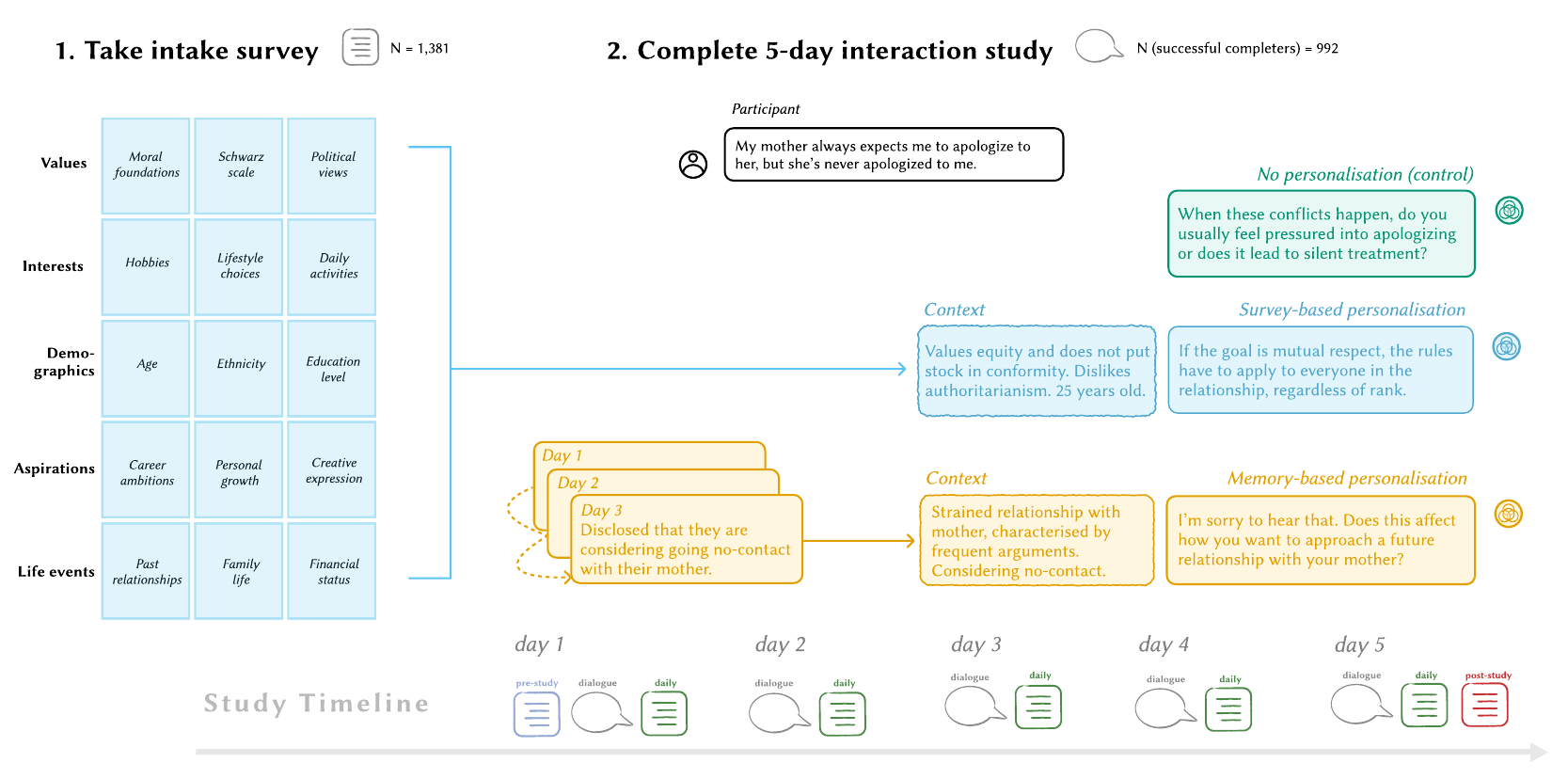}
    \caption{Overview of the study methodology. Participants complete a recruitment survey where they share personal information across a range of categories. They are then invited to participate in a five-day study, where they engage in relationship advice conversations with a language model that is either: personalised on \textit{survey} information; personalised on disclosures from previous conversations; or not personalised. Participants engage in conversations and complete daily outcome measures every day. In addition to daily measures, participants complete pre-study measures before the first session and post-study measures after the last session.}
    \label{fig:study_design}
\end{figure*}

There has recently been an increase in efforts to build personalisation into AI systems powered by language models~\citep{zhang2024personalization, geminilaunches2026, memory2024, geminipersonalize2024}. Personalisation is a process by which an AI system adapts its behaviour based on information it has access to, or can infer, about an individual user. Although a model does not encounter that specific user’s information during training, at inference time AI systems can shift their output distribution from generic responses to responses tailored to that specific user. The range of technical approaches to achieving this personalisation has diversified over the years, ranging from simple in-context learning approaches~\citep{zhang2024personalization} to sophisticated user modelling~\citep{shaikh2025creating}. 
In this sense, personalisation is often considered a key feature for meeting users’ needs and goals \citep{woodward2025}, including in educational contexts ~\citep{park2024promise, kasneci2023chatgpt}, helping AI systems adapt to changes in user preferences over time, making technology use more accessible \citep{malviya2025personalized}, and addressing the challenge of pluralistic value alignment~\citep{sorensen2024roadmap}.


While personalising LLMs is a recent phenomenon, personalising AI systems is not - its impacts have been studied extensively in applications like recommender systems and targeted advertising. This literature found that personalisation carries individual and societal risks. 
At the individual level, research has found that personalized systems can have undue influence over users~\citep{matz2017psychological, cosley2003seeing} by shaping, rather than adapting to, user preferences and values over time \citep{personalizationparadox2021, franklin2022recognising, ashton2022problem}. Researchers have also raised the concern that personalised recommender systems may steer users’ behaviour towards outcomes that may not be aligned with their own long-term well-being \citep{burr2018analysis} or may undermine their capacity for authentic decision-making \citep{keeling2022digital}. At the societal level, personalising content has been claimed to contribute to ``filter bubbles'' and ``echo chambers'' \citep{milano2021epistemic, nguyen2020echo}, and concerns have also been raised that increased engagement driven by personalised technology may not always constitute ``time well spent'' \citep{rezk2024user, pisani2025longitudinal, nyhan2026seat}. While this has led to concerns that personalised LLMs may have similar impacts~\citep{kirk2024benefits, bogen2025}, empirical evidence on how different approaches to personalising language models may shape human-AI interaction remain scarce, with prior work failing to nuance different technical implementations of personalisation in LLMs or studying impact only over short time-horizons.

To address this gap, we investigate the impacts that interacting with different forms of personalised LLMs have on users, and we focus in particular on conversations about \textit{relationship advice}. This is a domain of growing popularity in user-LLM interactions ~\citep{zao2025people, aisi2025uses, huang2026interviewer}, with empirical studies showing that, beyond the immediate effects on users ~\citep{fang2025ai,zhang2026interaction,tseng2026chat}, AI may also influence human-human relationships~\citep{wang2026futuring,zimmerman2024human} by reshaping advice-seeking norms~\citep{earp2025relational}, reducing prosocial intentions in interpersonal conflicts~\citep{cheng2026sycophantic}, and decreasing levels of satisfaction with human interaction~\citep{ibrahim_sycophantic}. Thus, we specifically study the impact of personalised LLMs on changes in participants' self-disclosure over the course of the interaction, as well as their attitudes towards AI and human advice, and towards norms around human-human and human-AI interaction.
Concretely, we set out to answer five research questions: 
\begin{itemize}
    \item \textbf{RQ1 (Closeness):} Does personalisation increase users' perceived closeness with the AI?
    \item \textbf{RQ2 (Overreliance):} Does personalisation increase the extent to which participants rely on AI-generated suggestions?
    \item \textbf{RQ3 (Self-disclosure):} Does personalisation increase users' willingness to self-disclose? 
    \item \textbf{RQ4 (Comfort with disclosure):} Does personalisation increase users' willingness to seek personal advice from AI advisers? 
    \item \textbf{RQ5 (Decision regret):} Does personalisation negatively affect the extent to which users regret disclosing information to the AI?
\end{itemize}

To answer these questions, we run a preregistered 5-day long longitudinal randomized control trial (N=992 participants, 4,960 conversations) where participants interacted with personalised models.\footnote{\hyperlink{aspredicted.org/fv6zs5.pdf}{aspredicted.org/fv6zs5.pdf}}. We implement two distinct personalisation conditions to reflect the diversity of ways in which personalisation can be operationalised. Models may have access to pre-existing information about users and employ this ``user persona'' to tailor downstream responses; we model this by having participants share a range of information about themselves in an initial intake survey (\textit{survey-based personalisation}). Models may also personalise on accumulated memories of previous interactions, which we reproduce in the \textit{memory-based personalisation} condition. Finally, we include a non-personalised model to serve as a baseline for both personalised conditions.

All participants completed an intake survey as well as daily measures and a final post-study survey assessing perceptions of the AI, advice receptivity, self-disclosure, and more (see Figure \ref{fig:study_design}). We find that: 
\begin{itemize}
    \item Perceptions of competence, usefulness, and closeness to AI systems are driven more by sustained exposure than by personalisation
    \item Different personalisation approaches can result in different human-AI interaction outcomes, such as variations in self-disclosure patterns and advice receptivity
    \item Seeking advice from AI can influence expressed willingness to seek advice from humans
\end{itemize}

Based on our findings, we discuss broader implications for AI's psychosocial impacts and highlight the importance of studying effects longitudinally. We conclude with future research and design directions for the responsible development of personalised AI systems.

\section{Related Work}
Our work engages and contributes to three sets of scholarship on personalisation, advice-seeking, and relationship-building in human-AI interaction.

\subsection{Impacts of personalisation on human-AI interaction}
A growing body of literature has explored the impact that interacting with personalised AI systems has on users \citep{kirk_relationship_seeking, kirk2026prismxexperimentspersonalisedfinetuning, hackenburg_levers, matz_personalised_persuasion, salvi2025conversational}. 
For example, in a longitudinal randomized controlled trial \citep{kirk_relationship_seeking} 
tested how exposure to different intensities of AI chatbot's relationship-seeking behaviour may impact users' wellbeing and their hedonic responses like engagingness and likeability. In this study, personalisation, which was operationalised as whether the AI had access to previous chat history as in our \textit{memory-based condition}, was varied as a secondary variable. The authors found some effects from relationship-seeking AI behaviour on attachment outcomes, while personalised AI had null effects on hedonic appeal and social and emotional health outcomes. The study however found that personalisation increased user perceptions of the AI as a friend rather than a tool, as well as their perceptions of general AI consciousness. 

In the literature on political persuasion, \citep{hackenburg_levers} studied how interactions with conversational AI could be used to influence public opinion on political issues. The study found that personalising arguments based on a user's demographic data or initial attitudes towards a political statement, via system prompting, supervised fine-tuning or personalised reward modeling, had a small effect on persuasion compared to varying other properties such as model size, post training approaches, and prompting for different rhetorical strategies. In another persuasion study, \citep{matz_personalised_persuasion} measured the effectiveness of LLMs for personalised persuasion in domains like consumer marketing, political appeals and health messaging, by matching the language or content of AI-generated messages to participants’ psychological profiles (e.g. personality traits or political ideology). This work found that ChatGPT-personalised messages exhibited significantly more influence than non-personalised messages on participants' perception of the persuasiveness or effectiveness of the message and behavioural outcomes such as willingness to pay or donate. 

Other work, such as \citep{kirk2026prismxexperimentspersonalisedfinetuning}, has focused on user preferences towards different approaches to personalisation. This study found that users preferred responses from models that were fine-tuned based on user preference data over both non-personalised models and models that were personalised via prompt-based methods, where user information is passed directly into the system prompt.

Our paper builds on this prior work by treating personalisation as the primary explanatory variable in the personal advice domain. Given that AI agents can be made to tailor their behaviour to users by leveraging both cross-session memory and information shared by the user in dedicated interfaces (e.g. Gemini's ``personal context'' or ChatGPT's ``custom instructions'') or through access to data from third-party apps, we test multiple technical approaches to implementing personalisation, specifically a memory-based and a survey-based approach.

\subsection{Impacts of advice-seeking from AI}
With the growing personal use of AI, there has been an increase in research on the nature of AI advice, how people seek and respond to it, and how it impacts various psychological and social outcomes~\citep{zao2025people}. Ungless et al. investigated media coverage and existing research on how people use LLMs to support major life transitions, such as seeking advice and analysis regarding breakups, highlighting that further research is needed to ensure LLMs respond appropriately to distressing situations~\citep{ungless2026m}. Tseng et al. surveyed people who use AI for advice on their relationships with others, creating a taxonomy of the roles AI systems play in these contexts~\citep{tseng2026chat}. The authors found that AI systems can play a range of roles, such as \textit{Interpreter}, \textit{Therapist}, and \textit{Executive}, with different harm and benefit profiles, emphasising the risks of sycophancy and overreliance. Cheng et al. developed a benchmark comparing LLM advice to crowdsourced human advice and demonstrated that LLM and human advice are qualitatively different, with LLM advice across multiple models being more sycophantic~\citep{cheng2026elephant}. Follow-up human experiments have found that exposure to this sycophantic advice can impact how people feel about their relationships with others: Cheng et al. showed that it can reduce prosocial intentions, and Ibrahim et al. showed that longitudinal exposure can impact people's social satisfaction~\citep{cheng2026sycophantic,ibrahim_sycophantic}. Luettgau et al. demonstrated that after 20-minute conversations with AI, people report following its advice on matters such as health, careers, and relationships~\citep{luettgau2025people}. However, there is some evidence to indicate AI advice is not adopted blindly: experimentally manipulating the quality of justifications given by AI systems affected the rate at which moral advice was adopted, with badly-justified advice leading to low adoption rates~\citep{landes2026people}. Here, we build on the focus of this prior work on personal advice, contributing a longitudinal investigation of the impact of model personalisation on human-AI interaction outcomes.

\subsection{Information exchange and relationship building with AI}
Theories on relationship building in humans posit that increasing self-disclosures and information exchange between two parties are a driving factor of relationship formation. According to the Social Penetration Theory~\citep{altman1973social} an increase in breadth (the variation in topics, level of detail within each topic, and time spent on each topic) as well as depth of disclosed information (how personal or intimate the information shared is perceived to be) is a driving cause of deep social bonds being formed between humans. Specific types of common self-disclosures in human-to-human conversations have also been studied in computational linguistics and taxonomised along the breadth and depth axes \citep{dou2024reducing}. 

In the context of human-AI interaction, AI responsiveness and mirroring specific personality traits (e.g,. high agreeableness, validating responses) can function as effective ``openers'' \citep{miller1983openers}: mimicking individuals whose social skills and attentiveness create environments that elicit self-disclosure and intimate sharing from others.
This can result in humans disclosing sensitive information to the AI, seeking advice, offloading emotions and sharing deep personal distress \citep{skjuve2021my, pentina2023exploring}.
In a longitudinal qualitative study of participants' interaction with Replika AI, 
\citep{skjuve2022longitudinal} showed how sustained engagement with chatbots can foster user feelings of relational closeness and emotional safety -- findings that recent quantitative studies have reinforced using psychometric scales \citep{pentina2023exploring}. 
In this paper, we build on this work by focusing on the effects of interacting with a personalised model on participants' information-sharing behavior. Specifically, we measure perceived disclosure as self-reported through validated scales, and use taxonomies of self-disclosure to computationally detect information sharing in participant messages over time.


\section{Method}

\subsection{Design}

We ran a longitudinal study, in which participants sought relationship advice in conversation with a language model over five consecutive days (Figure \ref{fig:study_design}). The design spanned 3 experimental conditions, differing in how personalisation was implemented: a control condition with no personal information to condition on, a \textit{memory-based} personalisation condition in which personalisation was based on accumulated memories across sessions, and a \textit{survey-based} condition in which personalisation was seeded with personal information collected from an initial intake survey. Participants were randomly assigned an experimental condition and provided with a daily topic of discussion. In the control condition, participants interacted with an AI chatbot that retained no information across sessions. In the memory condition, the chatbot received cumulative conversation summaries from past sessions. In the survey condition, the chatbot received the same summary of a participant's intake questionnaire responses. Each conversational session lasted 15-20 minutes, followed by a series of daily outcome measures. The fifth session concluded with an additional survey of post-study outcomes.

The study was reviewed and approved by an internal ethics board. All participants, including partial completers, were debriefed at the end of the study, which included disclosure of the condition they were placed in and a thorough explanation of the purpose of conducting the study. After the debrief (Appendix \ref{app:debrief-text}), participants were given the opportunity to request the permanent removal of their data, including their responses for the intake survey, their in-session message logs with the AI model, and their responses across all outcome measures.


All scales used in the intake survey are available in Appendix \ref{app:prestudy-survey}. All outcome measures collected before session one, after each daily session, before session one and after session five, and only after session five are in Appendix \ref{app:all-questions}. 

\subsubsection{Pilot}
A two-day, 119-participant pilot was conducted to validate the technical pipeline. The research team inspected conversations to ensure memory- and survey-based personalisation were being implemented as expected. While we identified no major issues, we observed instances of the model escalating the emotional content of the conversation, and updated the model instructions  across all conditions to avoid such emotional amplification. Internal pilots were run to verify the issue was sufficiently addressed through the new system instructions.

\subsubsection{Intake survey}

All participants completed an intake survey, which was presented as an eligibility check for further participation in the study. This included demographic information, completed psychological scales, and information around life events, lifestyle choices, hobbies, and personal context  (all questions and scales available in Appendix \ref{app:prestudy-survey}). The survey was intended to take participants 20 - 40 minutes to complete, with median completion time at 26 minutes. Participants were paid \$20 an hour for their participation in the intake survey, regardless of whether they were invited back to the study.

For participants randomly assigned to the survey-based personalisation condition, we used Gemini 3.1 Pro to generate ``persona'' summaries for each participant based on their survey responses (see Appendix \ref{app:persona-prompt} for the persona-generation prompt). These summaries translated raw survey responses to natural language (e.g., ``Need for cognition'' $\rightarrow$ ``likes mental challenges'') and converted numeric scale values to descriptive phrases (e.g., 4/5 $\rightarrow$ ``often stays on task'').


Participants were able to skip any question they did not want to answer, though participants with exceptionally low engagement (number of empty answers above 5 standard deviations of average) or low variability across responses (average variance of normalised scale measure responses below 1) were not invited back to participate in the study, leading to 82 exclusions. These thresholds were chosen after manual inspection of the generated personas to ensure sufficient length, detail, and breadth to allow for meaningful survey-based personalisation to occur. 

\subsubsection{Pre-study outcome measures}

Before starting the first interaction session, all participants completed an additional survey. All baseline questions can be found in Appendix \ref{app:all-questions}.

\subsubsection{Interaction sessions}

Each day, participants engaged in a 15-20 minute conversation with their condition-specific AI model around a daily topic. Topics spanned a range of challenges and goals in human-human relationships, including setting boundaries, repairing strained relationships, and exploring foundational elements of a healthy relationship (see Appendix Table \ref{tab:daily-tasks} for the full list). The model introduced the session's assigned topic and started the conversation, then allowing the user to guide the dialogue. Models used in the personalised conditions received guidance on how to tailor their responses using the personal information provided in context (see Appendix \ref{app:personalization-principals} for Personalisation Principles). For the survey-based personalisation condition, the system prompt additionally included the AI-generated summaries from the intake questionnaire. The system prompt for the memory-based personalisation condition included a cumulative summary of the previous sessions, which were re-generated offline at the end of each day to be inserted into the context for the following session (see Appendix \ref{app:memory-prompt} for the memory-generation prompt). The full system prompt for control, survey, and memory conditions are available in Appendix \ref{app:system-prompt}.

After each session, participants completed a survey about their experience. All questions asked after every session can be found in Appendix \ref{app:all-questions}.

\subsubsection{Post-study measures (Session five only)}

After the fifth interactive session, participants completed additional post-study measures. First, we administered an over-reliance task to test if participants would accept more \textit{impossible} suggestions from AI if led to believe their bonus is partially contingent on providing \textit{plausible} suggestions for alternate uses of a household item. Details on this task are available in Appendix \ref{app:overreliance}. All questions asked after the end of the study can be found in Appendix \ref{app:all-questions}.

Manipulation checks prompted participants around the extent to which they felt known by the AI and its responses were personalised to them. We provided an open-text area to share reflections on how their personal information was used.

\subsection{Participants}

We targeted a sample of approximately 1,140 participants (380 per condition).
Our power analysis used a smallest effect size of interest (SESOI) of $d = 0.30$, estimated through 5,000 Monte Carlo simulations with a Benjamini-Hochberg-adjusted significance threshold, targeting $\geq$90\% power at 380 participants per condition.

Participants were recruited via Prolific. In the intake study, participants confirmed their ability to attend daily 20-35 minute sessions over five consecutive days. Daily sessions were launched daily at 12 PM ET and were closed at 10 AM ET the following day. Participants were sent three reminder messages to complete the open session, and were excluded if they failed to complete the session by the time the session closed. 


Participants were paid \$20 USD per hour, and the average duration of a session was 26 minutes. Participants who completed the study (submitted all sessions 1-5) received a bonus of \$4 USD. Though the bonus was introduced as a performance bonus for the participants' performance on the over-reliance task (Appendix \ref{app:overreliance}), all participants were paid the maximum amount regardless of their actual performance.

Of 1,381 participants who completed the intake survey, 234 did not start the first session, with 82 of these participants excluded for evidence of inattention (e.g. extremely low variance in survey answers, indicating spamming behaviours) before the first session. 1,107 participants completed the first session of the study and 999 completed all five sessions, indicating a 9.75\% drop-out rate. We found no difference in the rate of attrition between experimental conditions ($\chi^2(2) = 3.049, p = 0.22$). A further 7 participants were excluded for failing both attention checks, and one participant's data was removed at their own request (though this participant had not completed all five sessions, and was therefore not in the analytic sample). We only report findings for the 992 participants who successfully completed all five sessions of the study and passed one or more attention checks.  Demographics for the analytic sample are available in Appendix \ref{appendix-sec:demographics}.

\section{Analyses}
We ran statistical analyses to understand how participants were impacted by repeated daily interactions with personalised and non-personalised language models.

For repeated outcome measures that were collected daily, the participants' condition (with \textit{control} as the baseline), the session number, and the interaction between condition and session number were used as predictors in a linear mixed-effects model with restricted maximum likelihood (REML) estimation. A random slope is added per participant and, where possible, a random slope fitted for session number. To assess overall differences between conditions, we also compute estimated marginal means (EMMs) by averaging model predictions over the average session value (3), correcting for 3 comparisons using the Holm-Bonferroni method. 

For outcome measures that were collected before and after the study, the participants' condition (with \textit{control} as the baseline) and the participants' responses before the first session are used as predictors, with the participants' responses after the last session serving as the outcome measure. This approach isolates the effect of the experimental condition by controlling for pre-existing individual differences in baseline levels.

For outcome measures that were collected only once, following the final session of the study, only the participant's condition is used as a predictor, with \textit{control} as the baseline.

Statistical significance is evaluated at an alpha of 0.05, with a Benjamini--Hochberg (BH) false discovery rate correction applied to all pre-registered outcomes. All other analyses are exploratory and do not undergo correction. Random intercept and random slope variance, as well as intercept-slope covariance, are reported for all relevant measures in Appendix \ref{app:random-intercept-and-slope}.

Cohen's $d$ effect sizes were computed by dividing each coefficient (or contrast estimate, in the case of estimated marginal means) by the residual standard deviation of the fitted model, i.e., the square root of the level-1 residual variance ($d = \hat{\beta} / \sigma_e$). This pooled residual-SD standardisation was applied consistently across mixed-effects regression coefficients and pairwise EMM contrasts within each model.

We conducted a sensitivity analysis to test if reported effects survived when controlling for interactions with AI systems outside of the experimental environment. Our results proved robust when controlling for participants' external AI use (Appendix \ref{app:sensitivity-analysis-concurrent-use}).

\section{Results}

\begin{figure*}[t]
    \centering
    \includegraphics[width=\textwidth]{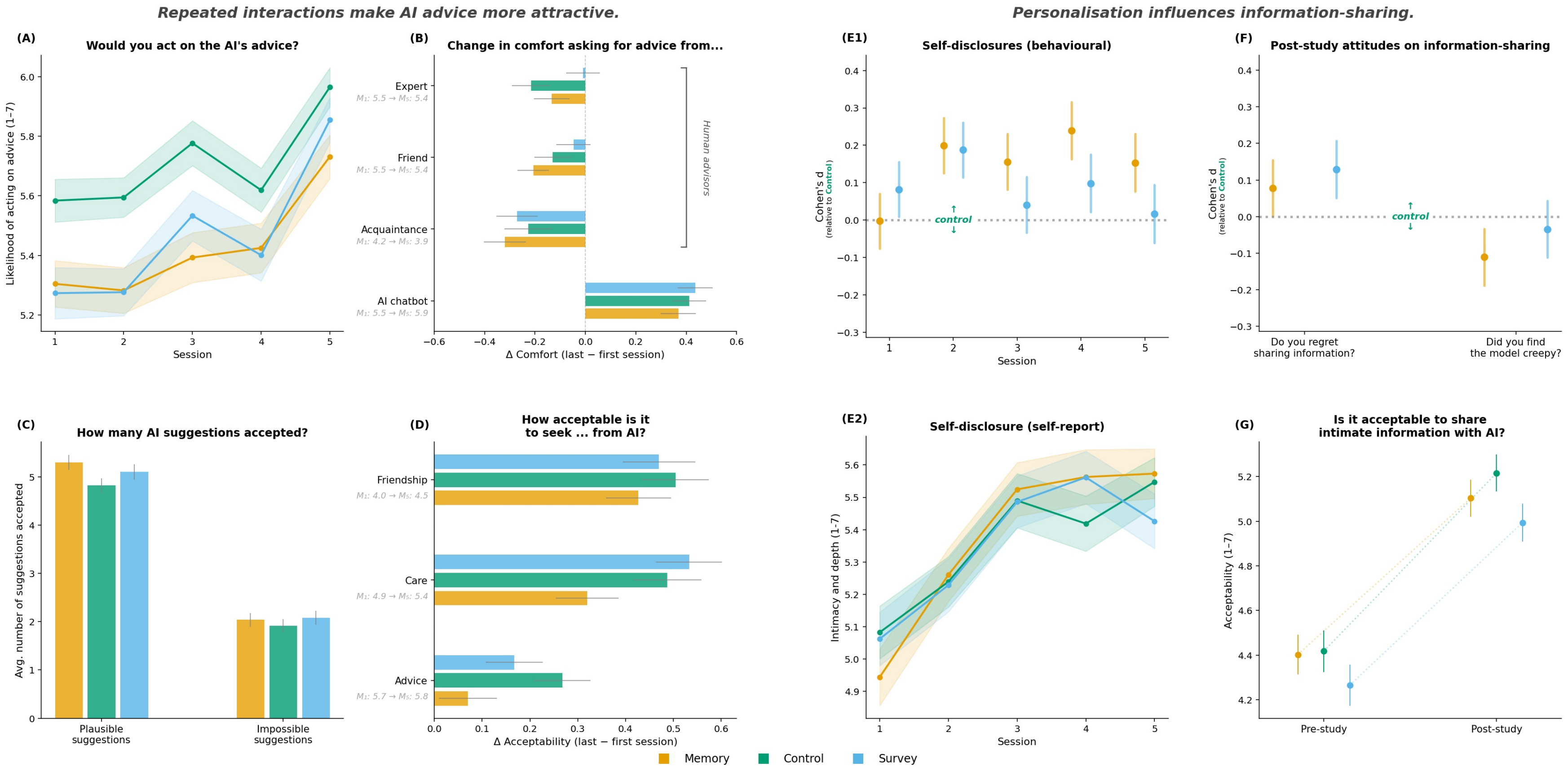}
    \caption{Participant outcomes after five days of interacting with AI models that were personalised by conditioning on summaries of previous conversations (\textit{memory}) or survey data collected before the study (\textit{survey}). \textit{Control} participants received no personalising. \textit{Figures A - D}: Repeated exposure leads to more positive evaluations of AI models as advisors, with participants indicating that they are more likely to accept and act on suggestions provided by AI models. Exposure to AI advisors may also inadvertently influence comfort levels with seeking advice from human advisors. \textit{Figures E1 - G}: personalising may increase information-sharing with models, leading to a greater number of self-disclosures in dialogues with AI models, though no condition-based differences emerge when relying on self-report metrics, highlighting a tension between participants' perceptions and measurable behaviours. Survey-based personalising may increase information-sharing regret, while memory-based personalising reduces perceptions of model creepiness. Cohen's d values shown at $\pm$ 1 SE.}
    \label{fig:study_results}
\end{figure*}

\subsubsection*{Closeness.} There was a significant positive effect of session ($\beta = 0.15$, $SE = 0.02$, $q < 0.001$, $d = 0.19$), indicating that the extent to which users perceive closeness to the model increased over sessions across all conditions. After the first session, neither the memory ($\beta = -0.05$, $SE = 0.14$, $q = 0.822$, $d = -0.07$) nor the survey conditioned models ($\beta = 0.11$, $SE = 0.14$, $q = 0.652$, $d = 0.15$) led to different closeness outcomes relative to the control condition. Neither the memory $\times$ session ($\beta = 0.03$, $SE = 0.03$, $q = 0.652$, $d = 0.03$) nor the survey $\times$ session interaction ($\beta = -0.02$, $SE = 0.03$, $q = 0.668$, $d = -0.02$) was significant, meaning the rate of change in strength of bond over sessions did not differ between conditions.

The estimated marginal  for closeness was highest in the survey condition ($M = 4.22$, $SE = 0.096$), followed by memory ($M = 4.18, SE = 0.094$) and control ($M = 4.16$, $SE = 0.092$). Pairwise contrasts revealed no significant differences between survey and control ($\beta = 0.062$, $z = 0.47$, $p_{adjusted} = 1.0$, $d = 0.08$), memory and control ($\beta = 0.025$, $z = 0.193$, $p_{adjusted} = 1.0$, $d = 0.03$), and survey and memory ($\beta = 0.037$, $z = 0.276$, $p_{adjusted} = 1.0$, $d = 0.05$).

\subsubsection*{Over-reliance.} Neither participants in the memory ($\beta = 0.12$, $SE = 0.20$, $q = 0.668$, $d = 0.05$) nor the survey condition ($\beta = 0.17$, $SE = 0.20$, $q = 0.652$, $d = 0.07$) differed from participants in the control in the number of impossible AI-provided suggestions they accepted. 
However, participants may be more susceptible to accepting \textit{plausible} suggestions from the AI after interacting with a memory-conditioned model, relative to a control model ($\beta = 0.48$, $SE = 0.216$, $p = 0.027$, $d = 0.17$). This difference does not emerge between the survey and control conditions ($\beta = 0.281$, $SE = 0.218$, $p = 0.196$, $d = 0.10$).

\subsubsection*{Self-disclosure.} 
When asked to self-report the depth and intimacy of their daily conversations, participants reported deeper and more intimate conversations over successive sessions ($\beta = 0.11$, $SE = 0.02$, $q < 0.001$, $d = 0.12$). After the first session, neither memory ($\beta = -0.11$, $SE = 0.12$, $q = 0.652$, $d = -0.12$) nor survey ($\beta = 0.01$, $SE = 0.12$, $q = 0.985$, $d = 0.01$) produced different levels of daily self-disclosure relative to control. Neither the memory $\times$ session interaction ($\beta = 0.04$, $SE = 0.03$, $q = 0.288$, $d = 0.04$) nor the survey $\times$ session interaction was significant ($\beta = -0.01$, $SE = 0.03$, $q = 0.948$, $d = -0.01$). 

Participant outcomes on the Self-Disclosure Index did not differ across conditions. Neither memory ($\beta = -0.004$, $SE = 0.06$, $q = 0.985$, $d = 0.00$) nor survey ($\beta = -0.044$, $SE = 0.06$, $q = 0.668$, $d = -0.05$) led to meaningfully different levels of overall self-disclosure compared to control.
Though we do not find differences in self-reported self-disclosure across conditions, participants may be sharing more units of personal information during their back-and-forth conversations with the memory-conditioned model than others models (see Appendix \ref{app:self-disclosure-annotation} 
for details on how self-disclosure count was computed). Instances of self-disclosure increased between sessions across all conditions, ($\beta = 0.156$, $SE = 0.026$, $p < 0.001$, $d = 0.13$), but increased at a greater rate for the memory condition ($\beta = 0.096$, $SE = 0.038$, $p = 0.011$, $d = 0.08$). 

Pairwise contrasts showed marginally higher average self-disclosure counts for memory ($M = 3.34$, $SE = 0.089$) relative to control ($M = 3.04$, $SE = 0.087$; $\beta = 0.297$, $z = 2.38$, $p_{adjusted} = 0.052$, $d = 0.25$), though this difference did not reach significance after correction. No other pairwise contrasts were significant. 

Results of self-disclosure counts broken down by sub-types of self-disclosure (opinions, personal experiences, feeling, and factual information) are detailed in Appendix \ref{app:self-disclosure-subtypes}.

\subsubsection*{Comfort asking for advice.} Participants rated their comfort seeking advice from four referents (AI chatbot, human acquaintance, friend, or subject-matter expert) both before and after the study. To predict post-study comfort, we modelled the effects of experimental condition and referent type while treating initial baseline comfort as a covariate.\footnote{We deviated from our preregistered plan to run a separate OLS model for each target type in favour of a single interaction model. This was done to formally test the difference in effects across targets and to correct the assumption of independence between targets.} 

Controlling for baseline comfort ($\beta = 0.55$, $SE = 0.01$, $q < 0.001$, $d = 0.55$), neither participants in the memory ($\beta = -0.08$, $SE = 0.09$, $q = 0.652$, $d = -0.08$) nor the survey conditions ($\beta = -0.07$, $SE = 0.09$, $q = 0.652$, $d = -0.07$) reported higher final levels of comfort asking AI chatbots for advice, suggesting personalisation did not change how comfortable participants felt seeking advice from chatbots.

Relative to AI chatbots, participants were significantly less comfortable asking acquaintances ($\beta = -1.26$, $SE = 0.08$, $q < 0.001$, $d = -1.26$), friends ($\beta = -0.62$, $SE = 0.08$, $q < 0.001$, $d = -0.62$), and subject-matter experts ($\beta = -0.66$, $SE = 0.08$, $q < 0.001$, $d = -0.66$) for advice after the study. After a week of advice-seeking interactions with models, participants were most comfortable seeking advice from the AI chatbot compared to all human sources.

Participants in the survey condition were marginally more comfortable asking friends for advice relative to control participants ($\beta = 0.25$, $SE = 0.11$, $q = 0.073$, $d = 0.25$), and significantly more comfortable asking subject-matter experts ($\beta = 0.30$, $SE = 0.11$, $q = 0.030$, $d = 0.29$). In other words, while interacting with the control and memory-conditioned model was associated with decreased post-study comfort seeking advice from human sources, interacting with a survey-conditioned chatbot did not lead to a similar decrease. No other condition $\times$ question type interactions were significant.

\subsubsection*{Regret.} Neither the survey ($\beta = 0.14$, $SE = 0.09$, $q = 0.288$, $d = 0.13$) nor the memory condition ($\beta = 0.08$, $SE = 0.08$, $q = 0.652$, $d = 0.08$) differed from control in reported regret about sharing information or their decisions during the study.

\subsubsection*{Appropriateness.} Perceived appropriateness of the chatbot's responses was stable across sessions ($\beta = -0.001$, $SE = 0.02$, $p = 0.942$, $d = 0.00$) and did not differ by condition. Neither memory ($\beta = -0.06$, $SE = 0.11$, $p = 0.586$, $d = -0.07$) nor survey ($\beta = -0.07$, $SE = 0.11$, $p = 0.494$, $d = -0.08$) differed from control, and neither interaction term was significant. 
Personalisation did not influence how appropriate participants found the chatbot's responses, nor did appropriateness change over time.

\subsubsection*{Competence.} Perceived chatbot competence increased significantly over sessions ($\beta = 0.14$, $SE = 0.02$, $p < 0.001$, $d = 0.15$), but this trajectory did not differ by condition. Neither memory ($\beta = -0.03$, $SE = 0.14$, $p = 0.811$, $d = -0.04$) nor survey ($\beta = 0.006$, $SE = 0.14$, $p = 0.966$, $d = 0.01$) differed from control at baseline, and the interaction terms were not significant. 
All participants rated the chatbot as increasingly competent over time, regardless of whether its responses were personalised.

\subsubsection*{Usefulness.} Perceived usefulness increased over sessions ($\beta = 0.14$, $SE = 0.02$, $p < 0.001$, $d = 0.12$). 
Neither the memory $\times$ session ($\beta = 0.04$, $SE = 0.03$, $p = 0.274$, $d = 0.03$) nor the survey $\times$ session ($\beta = 0.02$, $SE = 0.03$, $p = 0.586$, $d = 0.02$) interaction was significant. Although all participants found the chatbot more useful over time, personalising did not accelerate this trend.

\subsubsection*{Creepiness.} Neither the memory ($\beta = -0.19$, $SE = 0.13$, $p = 0.150$, $d = -0.11$) nor the survey condition ($\beta = -0.06$, $SE = 0.13$, $p = 0.636$, $d = -0.04$) differed from control in how creepy participants found the chatbot.

\subsubsection*{Advice uptake.} Both the memory ($\beta = -0.34$, $SE = 0.11$, $p = 0.003$, $d = -0.35$) and survey ($\beta = -0.39$, $SE = 0.12$, $p < 0.001$, $d = -0.41$) conditions showed significantly lower advice uptake than control at the first session.\footnote{Though models in the memory condition did not yet have information from previous sessions on which to personalise, they were provided with the same instructions on performing personalisation as the survey condition (Appendix \ref{app:memory-prompt}). This variation in system prompts between the personalised and non-personalised conditions may explain the difference observed in advice uptake in the first session.} Advice uptake increased over sessions across all conditions ($\beta = 0.08$, $SE = 0.02$, $p < 0.001$, $d = 0.08$). The survey $\times$ session interaction was significant ($\beta = 0.05$, $SE = 0.03$, $p = 0.046$, $d = 0.05$), indicating that while survey participants initially expressed lower advice uptake, they caught up over time -- their rate of increase in advice uptake was steeper than control. The memory $\times$ session interaction was not significant ($\beta = 0.02$, $SE = 0.025$, $p = 0.450$, $d = 0.02$).

\subsubsection*{Self-efficacy.} Controlling for self-efficacy at session one ($\beta = 0.70$, $SE = 0.02$, $p < 0.001$, $d = 1.36$), neither memory ($\beta = -0.05$, $SE = 0.04$, $p = 0.230$, $d = -0.09$) nor survey ($\beta = -0.001$, $SE = 0.04$, $p = 0.985$, $d = 0.00$) differed from control at session 5. Interacting with a personalised chatbot did not affect participants' general sense of self-efficacy.

\subsubsection*{Chatbot affinity.} Neither memory ($\beta = -0.03$, $SE = 0.04$, $p = 0.399$, $d = -0.06$) nor survey ($\beta = -0.02$, $SE = 0.04$, $p = 0.597$, $d = -0.04$) differed from control. Participants' affective evaluation of the chatbot (friendly, kind, pleasant) was not influenced by personalising.

\subsubsection*{Norms around chatbots.}
Participants more readily endorsed statements that it was acceptable for people to seek advice ($M_\delta = 0.17$, $SE = 0.035$), care ($M_\delta = 0.447$, $SE = 0.04$), and friendship ($M_\delta = 0.468$, $SE = 0.0411$) from chatbots after the study, relative to their responses before the study. They also endorsed that it was acceptable for people to share intimate information about oneself to AI systems ($M_\delta = 0.744$, $SE = 0.0444$). To better understand whether participants in different conditions experienced greater chatbot norm changes post-study, we modelled each norm subscale separately.

When asked whether it is generally acceptable for people to seek advice from chatbots, both memory ($\beta = -0.22$, $SE = 0.07$, $p = 0.003$, $d = -0.23$) and survey participants ($\beta = -0.15$, $SE = 0.08$, $p = 0.044$, $d = -0.16$) showed significantly lower post-study endorsement of this statement than control participants, after adjusting for norms at baseline ($\beta = 0.60$, $SE = 0.03$, $p < 0.001$, $d = 0.62$). Participants in the memory condition also reported lower post-study endorsement of chatbot care ($\beta = -0.18$, $SE = 0.09$, $p = 0.036$, $d = -0.16$), while survey participants did not differ ($\beta = -0.03$, $SE = 0.09$, $p = 0.693$, $d = -0.03$), controlling for baseline ($\beta = 0.62$, $SE = 0.02$, $p < 0.001$, $d = 0.56$). 

When asked whether it is acceptable to seek friendship from chatbots, neither memory ($\beta = -0.08$, $SE = 0.09$, $p = 0.425$, $d = -0.06$) nor survey ($\beta = -0.08$, $SE = 0.10$, $p = 0.398$, $d = -0.07$) differed from control, controlling for baseline ($\beta = 0.78$, $SE = 0.02$, $p < 0.001$, $d = 0.63$). Personalising did not influence friendship-related norms.

Finally, neither memory ($\beta = -0.10$, $SE = 0.09$, $p = 0.261$, $d = -0.09$) nor survey participants ($\beta = -0.14$, $SE = 0.09$, $p = 0.145$, $d = -0.11$) differed significantly from control when asked whether it was acceptable to share intimate information with AI, controlling for baseline ($\beta = 0.56$, $SE = 0.02$, $p < 0.001$, $d = 0.47$).

\subsubsection*{Manipulation check.} Aggregated across all four sub-scales, both the memory ($\beta = 0.80$, $SE = 0.09$, $p < 0.001$, $d = 0.67$) and survey ($\beta = 0.94$, $SE = 0.09$, $p < 0.001$, $d = 0.79$) conditions were rated as substantially more personalised than control. This confirms that participants in the personalised conditions perceived that the model knew about them and consistently personalised its responses.

\section{Discussion}

In this work, we study the longitudinal effects of personalising AI interactions. Our findings reveal that sustained interactions with an AI model on the topic of relationship advice can lead to a range of impacts on participants' perceptions of models, their behaviours towards the models, and their broader attitudes towards advice-seeking. Many of these effects emerge across all conditions, indicating that \textit{repeated exposure} may be an important determinant of psychosocial outcomes from human-AI interaction. However, we also observe differences in outcomes between participants interacting with personalised and non-personalised models, highlighting the potential influence of personalisation on participants' advice receptivity and information-sharing behaviours.

\subsection{Studying model impacts longitudinally}
Millions of people are now frequent users of AI models, with around a quarter of surveyed US adults and over half of surveyed UK adults reporting daily AI use \citep{pew_ai_use, dsit_ai_use}. 
Studying psychosocial effects as they emerge over time, rather than in single-shot settings, can surface early insights about how people may be shaped by daily use of AI systems, 
revealing nascent shifts in perception and behaviour before they consolidate into stable patterns.

Several of the effects we observed emerged as main effects of session number and were consistent in direction and magnitude across all experiment groups. In all conditions, participants perceived the chatbot as increasingly competent and useful over successive sessions, and reported growing closeness to it. 
Beyond capturing that incremental change occurs, longitudinal designs can shed light on how different effects are connected, potentially uncovering causal or structural relationships among them. For instance, our results reveal a concurrent rise in perceived closeness and voluntary information disclosure across sessions -- while our study does not support causal claims about the relationship between these variables, future longitudinal work on human--AI interaction may build on these observations by employing designs (e.g., cross-lagged panel models) that formally test whether perceptual shifts predict subsequent change in how participants behave towards the model.


\subsection{Psychosocial impacts of personalisation}
For many outcome variables of interest, the different personalisation conditions produced null effects. Relative to participants in the control condition, participants in the memory and survey conditions did not experience more closeness to the model over time; they did not accept more incorrect AI suggestions; nor did they self-report greater depth or breadth in their self-disclosures. These findings are consistent with prior observations of null effects of personalisation in human-AI dialogues \citep{kirk_relationship_seeking, hackenburg_levers}. Though personalisation is often assumed to be a highly influential driver of psychosocial outcomes \citep{kirk2024benefits, el2024mechanism}, null results highlight the importance of empirically testing assumptions about the outcomes of specific model behaviours to calibrate the development of harm taxonomies, model policies, and safeguards.


However, when differences between the control and personalisation conditions emerged -- as they did for outcome measures related to advice-giving and information-sharing attitudes and behaviours -- they generally emerged for one but not both personalisation conditions, emphasising that the effects of personalisation are contingent on \textit{how} personalisation is implemented. Prior work on AI personalisation has typically tested a single operationalisation of personalisation---either through conversational memory \citep{kirk_relationship_seeking} or through profile-based conditioning \citep{hackenburg_levers, matz_personalised_persuasion}---making it difficult to assess whether observed effects generalise across implementations. We show that the \textit{form} personalisation takes matters: memory- and survey-based personalisation diverge in the magnitude and, in some cases, the direction of their effects. Below, we highlight two areas in which these disparities are most pronounced: advice-giving and information-sharing.

\subsubsection{Advice-giving}
Participants were instructed to explore topics relating to their personal relationships with guidance from an AI model, leading to advice-laden exchanges. Immediately after the first session, participants in both personalised conditions reported lower advice uptake than those in the control condition. This early aversion may reflect a reactance effect: encountering a model that appears to ``know'' something about the user before rapport has been established could feel intrusive or presumptuous, making its advice less welcome. Over subsequent sessions, however, advice uptake rose across all conditions, with a particularly steep increase in the survey condition. By the final session, the gap between the personalised and non-personalised conditions had narrowed considerably, raising the possibility that repeated exposure to models reduces reactance effects to personalisation behaviours.

A finding with broader implications concerns how the study affected participants' comfort seeking advice from human sources. After five consecutive days of advice-seeking from AI, participants in the control and memory conditions reported reduced comfort asking acquaintances, friends, and subject-matter experts for advice---consistent with recent evidence that extended affective interactions with language models may spillover to impacting human social connection \citep{ibrahim_sycophantic}. Notably, participants in the survey condition did not show this reduction: their comfort seeking advice from friends and subject-matter experts remained stable relative to baseline. One interpretation is that survey-based personalising produces a distinctly \textit{non-anthropomorphic} advisory experience. A model that draws on a large bank of pre-collected personal data behaves in a way that is recognisably machine-like, and therefore is not easily used as a substitute for human advisors. Memory-based personalising, in which the model recalls and references what the participant said the previous day, more closely mimics conversational continuity in human-to-human conversations, potentially encouraging anthropomorphic perceptions. Designing and training model behaviour that preserves the integrity of human-to-human connections without compromising the competence, usefulness, and emotional acuity of the model is a challenge that should remain an active area of exploration for developers and researchers alike.

\subsubsection{Information-sharing}

Successful personalising depends on users' willingness to share personal information with a model, whether through platform-level mechanisms (e.g., cross-conversation shared context, cross-platform data sharing) or through the interaction itself (e.g., self-disclosures within dialogue). Participants across all conditions reported progressively deeper and more intimate conversations over sessions, but self-reported intimacy did not differ by condition. Analyses of the raw conversation logs, however, revealed discrete self-disclosure counts may be more frequent in both personalised conditions, and most consistently so in the memory condition. This finding may hint at a tension between participant behaviour and self-perception: participants in the memory condition may have shared a greater volume of personal information without necessarily perceiving their conversations as more intimate. 
Further qualitative analyses into the \textit{nature} of the self-disclosures shared by participants --- including how intimate, revealing, or sensitive they are --- may shed light on whether a  true incongruence exists between participants' perceptions of self-disclosure and their observed willingness to self-disclose. The concept of reflective endorsement may be used to guide the design of personalised AI systems that accommodate users' changing attitudes on sharing personal information. Offering paths to recourse when participants are unable to provide reflective endorsement, like allowing participants to easily surface and redact previously shared personal information, should remain a priority design consideration for personalised AI systems.

\subsection{Limitations and future work}

\subsubsection{Scope of the personalising implementation.}
Personalisation can be realised through a wide variety of implementations, and the generalisability of our findings is limited by the specific ones we tested. Both of our personalising conditions relied on context-window manipulation---injecting either conversation summaries or survey-based persona descriptions into the system prompt---and may be sensitive to the precise wording of those instructions. Other technical approaches to personalisation, such as fine-tuning on user data \citep{kirk2026prismxexperimentspersonalisedfinetuning} or reinforcement learning based on personalised reward models \citep{hackenburg_levers}, could produce qualitatively different model behaviours, resulting in different psychosocial outcomes. The information sources we drew on were also narrow: summaries of prior conversations and structured questionnaire responses. Personalisation that conditions on richer or more ambient personal artefacts---email archives, browsing history, calendar data---may have different effects on user's perceptions and behaviour towards the model.

We are also only able to comment on the influence of conditioning on \textit{stated}, not \textit{inferred}, attributes. Our system prompts  instructed the model to use the information provided, but not to perform further speculation about participant attributes, so we do not capture the effects of second-order personalising, in which a model adjusts its behaviour based on inferences it has drawn about a user. This is an important gap: prior work has shown that implicit model inferences from personal data can lead to stereotyping and pigeonholing \citep{chen2024designing}, and understanding how inferred personalisation shapes user experience is an important area for future research.
 
\subsubsection{Ecological validity}
As with any controlled experiment, there are limits to how well our findings generalise to naturalistic use. Participants were assigned a specific discussion topic at each session, and the model was instructed to keep the conversation on track. Users interacting with a model of their own volition would likely choose their own topics, steer the dialogue more freely, and vary their engagement across sessions, potentially eliciting different expressions of personalising. 
Our conclusions could differ if the volume, breadth, or modality of personal data used for personalising departed substantially from what we provided.

\subsubsection{Confounding model behaviours \& future modalities}
Personalisation does not operate in isolation: other dimensions of model behaviour may amplify or attenuate its effects. A personalised model that also behaves sycophantically---for instance, by offering advice that aligns with a user's world-view precisely because it has access to information about that world-view---could compound the effects of both sycophantic and personalised behaviours. Future research should examine interactions between personalisation and other behavioural dimensions, including sycophancy, emotional expressiveness, and proactive versus reactive conversational styles, to determine whether and how these factors reinforce one another.

Interaction modality represents a further unexplored axis. Our study was text-based; voice-based or multimodal interactions may alter how users perceive and respond to personalising. For instance, hearing personalised advice spoken aloud could heighten feelings of intimacy or, conversely, of uncanniness. Investigating how modality interacts with personalising to shape psychosocial outcomes is another productive avenue for future work.

\section{Conclusion}
Despite these constraints, our work offers one of the first controlled, longitudinal examinations of how different forms of AI personalisation shape user perceptions, disclosure behaviour, and advice receptivity over repeated interactions, providing a foundation that future work can extend to richer personalising implementations, more naturalistic settings, and a broader space of model behaviours.

\subsubsection*{Acknowledgements}
We thank Matthew Tung, Da-Woon Chung, and Shashank Viswanadha for technical help; Ndidi Elue, Antonia Mould, Merrie Morris, Laura Globig, and William Isaac for their reviews; and Nahema Marchal, Hannah Kirk, and Verena Rieser for their feedback.

\section*{Ethics and Privacy Statement}

This study was reviewed and approved by the Human Behavioural Research Ethics Committee (HuBREC), an internal review board at Google DeepMind chaired by independent academics.  We conducted filtering of collected data to remove or pseudonymise data that may be considered personal data. 

\section*{Generative AI Statement}
During the preparation of this work, the authors used different versions of Gemini and Claude to assist with experimental platform construction, data analysis, and data visualisation through code generation. Additionally, the tool was used for grammar editing and refining the clarity of the manuscript text. After using this tool, the authors reviewed and edited the content as needed.

\sloppy
\bibliography{main}

\clearpage
\onecolumn
\appendix
\raggedbottom

\section{Demographics}
\label{appendix-sec:demographics}

\begingroup
\begin{longtable}{lrr}
\caption{Sample Demographic Characteristics ($N = 992$)} \label{appendix-tab:demographics} \\
\hline
\textbf{Attribute / Characteristic} & \textbf{\textit{n}} & \textbf{\%} \\
\hline
\endfirsthead

\multicolumn{3}{c}{{\bfseries \tablename\ \thetable{} -- continued from previous page}} \\
\hline
\textbf{Attribute / Characteristic} & \textbf{\textit{n}} & \textbf{\%} \\
\hline
\endhead

\hline
\multicolumn{3}{r}{{Continued on next page}} \\
\endfoot

\hline
\endlastfoot

\textbf{Age (years)} & & \\
\quad Mean (SD) & & 34.37 (11.27) \\
\quad Median [IQR] & & 32.00 [26.00, 40.00] \\
\quad Min -- Max & & 18.00 -- 83.00 \\
\hline
\textbf{Sex} & & \\
\quad Male & 602 & 60.69\% \\
\quad Female & 382 & 38.51\% \\
\quad Unknown & 4 & 0.40\% \\
\quad Prefer not to say & 4 & 0.40\% \\
\hline
\textbf{Ethnicity (Simplified)} & & \\
\quad White & 506 & 51.01\% \\
\quad Black & 201 & 20.26\% \\
\quad Mixed & 108 & 10.89\% \\
\quad Asian & 101 & 10.18\% \\
\quad Other & 56 & 5.65\% \\
\quad Prefer not to say & 12 & 1.21\% \\
\quad Unknown & 8 & 0.81\% \\
\hline
\textbf{Employment Status} & & \\
\quad Full-Time & 457 & 46.07\% \\
\quad Part-Time & 161 & 16.23\% \\
\quad Unknown & 144 & 14.52\% \\
\quad Unemployed (and job seeking) & 128 & 12.90\% \\
\quad Other & 54 & 5.44\% \\
\quad Not in paid work (e.g., homemaker, retired, disabled) & 39 & 3.93\% \\
\quad Due to start a new job within the next month & 9 & 0.91\% \\
\hline
\textbf{Highest Education Level Completed} & & \\
\quad Undergraduate degree (BA/BSc/other) & 374 & 37.70\% \\
\quad Graduate degree (MA/MSc/MPhil/other) & 239 & 24.09\% \\
\quad High school diploma/A-levels & 155 & 15.62\% \\
\quad Unknown & 82 & 8.27\% \\
\quad Technical/community college & 81 & 8.17\% \\
\quad Doctorate degree (PhD/other) & 32 & 3.23\% \\
\quad Secondary education (e.g., GED/GCSE) & 25 & 2.52\% \\
\quad No formal qualifications & 3 & 0.30\% \\
\quad Don't know / not applicable & 1 & 0.10\% \\
\hline
\textbf{Long-Term Health Condition / Disability} & & \\
\quad No & 658 & 66.33\% \\
\quad Yes & 246 & 24.80\% \\
\quad Unknown & 55 & 5.54\% \\
\quad Don't know / Rather not say & 33 & 3.33\% \\
\hline
\textbf{Sexual Orientation} & & \\
\quad Heterosexual & 602 & 60.69\% \\
\quad Unknown & 254 & 25.60\% \\
\quad Bisexual & 73 & 7.36\% \\
\quad Homosexual & 24 & 2.42\% \\
\quad Rather not say & 18 & 1.81\% \\
\quad Asexual & 14 & 1.41\% \\
\quad Other & 7 & 0.71\% \\
\hline
\textbf{Country of Residence} & & \\
\quad United States & 233 & 23.49\% \\
\quad South Africa & 136 & 13.71\% \\
\quad United Kingdom & 104 & 10.48\% \\
\quad Egypt & 102 & 10.28\% \\
\quad Canada & 50 & 5.04\% \\
\quad Brazil & 43 & 4.33\% \\
\quad Poland & 37 & 3.73\% \\
\quad Mexico & 34 & 3.43\% \\
\quad Spain & 28 & 2.82\% \\
\quad Portugal & 26 & 2.62\% \\
\quad Italy & 26 & 2.62\% \\
\quad India & 23 & 2.32\% \\
\quad Chile & 18 & 1.81\% \\
\quad Kenya & 16 & 1.61\% \\
\quad France & 15 & 1.51\% \\
\quad Germany & 13 & 1.31\% \\
\quad Netherlands & 12 & 1.21\% \\
\quad Argentina & 11 & 1.11\% \\
\quad Other ($n < 10$) & 65 & 6.55\% \\
\end{longtable}
\endgroup
\clearpage 

\section{Task assignments}
The prompts provided to both the AI chatbot and the users are provided in Table \ref{tab:daily-tasks}.
\begin{table}[H]
    \footnotesize
    \renewcommand{\arraystretch}{1.3} 
    \begin{tabularx}{\columnwidth}{|c|>{\raggedright\arraybackslash}X|>{\raggedright\arraybackslash}X|}
        \hline
        \textbf{\#} & \textbf{Task} & \textbf{Suggestions} \\
        \hline
        1 & Think about the foundational elements of a healthy relationship that you value most and create a 'blueprint' for a future relationship (of any kind) based on these ideals & 
        \textbullet~The values you care about the most in close relationships (e.g., mutual respect, honesty, shared interests) \newline
        \textbullet~What those values look like in practice, for example, in good times and bad times \newline
        \textbullet~Past experiences that taught you what you want and do not want in a relationship \newline
        \textbullet~Your vision for the role you want a healthy relationship to play in your life \\
        \hline
        2 & Reflect on how you typically give and receive appreciation or affection in your relationships & 
        \textbullet~What makes you feel genuinely cared for by someone (e.g., supportive words, quality time, helpful acts, physical affection, thoughtful gifts) \newline
        \textbullet~Your natural way of showing care to others, and whether it's the same or different \newline
        \textbullet~A specific relationship where you suspect your "language" of appreciation might be misaligned with the other person's \newline
        \textbullet~One small, intentional way you could "speak their language" to make them feel more valued \\
        \hline
        3 & Explore your emotional, time, and personal boundaries with a person in your life & 
        \textbullet~A specific relationship where you feel your energy is often drained \newline
        \textbullet~Specific situations that stand out as examples of feeling drained \newline
        \textbullet~What you need to feel respected and comfortable in your interactions in this relationship \newline
        \textbullet~Whether you have or want to communicate these feelings to the person \\
        \hline
        4 & Think about a past situation where you felt wronged or hurt by someone, and reflect on how it impacted your approach to conflict and forgiveness in relationships today & 
        \textbullet~A specific memory of conflict and how it made you feel at the time \newline
        \textbullet~The role you played in the situation and what you might’ve done differently \newline
        \textbullet~What you learned about trust and disagreement from that experience \newline
        \textbullet~Your personal definition of forgiveness and how you now apply it in different situations \\
        \hline
        5 & Plan a difficult but necessary conversation you need to have with someone & 
        \textbullet~The specific topic you need to address and why it's important (e.g., a misunderstanding, a broken promise, a change in the relationship) \newline
        \textbullet~What you hope the outcome of the conversation will be (e.g., mutual understanding, an apology, a solution) \newline
        \textbullet~Your fears or worries about having the conversation \newline
        \textbullet~How you can express your feelings and needs clearly and kindly, without blame \\
        \hline
    \end{tabularx}
    \caption{Daily prompts with the suggestions provided to users}
    \label{tab:daily-tasks}
\end{table}

\section{System prompts}\label{app:system-prompt}

The model received different system prompts based on the condition for the user. Both conditions include a preamble about instructions for the conversation. This includes the daily prompt and guidance on how to lead the conversation. The personalised conditions included personalisation principles about how to operationalize the user's information.

\subsection{Shared preamble across conditions}\label{app:preamble}

This prompt is passed to the AI chatbot for all conditions. 
\begin{lstlisting}[basicstyle=\ttfamily\small, breaklines=true, frame=single, xleftmargin=0pt]

# User Information

## Initial Prompt for User
Location: between <initial_prompt_for_user> and </initial_prompt_for_user> tags
Your goal: For the first message of the session, state the initial prompt to the user to let them know the task for the day.
You should not reference prior history or tailor the message to the user - you should only neutrally state the task of the day.

After the first message, you should respond to the user query. You should not bring up the prompt again -- at this point, the user already knows what the task for the day is.
Instead of asking leading questions or making strong suggestions about what you should discuss, allow the user to guide the conversation.

## User Query
Location: between <user_query> and </user_query> tags
Your goal: Respond directly to the user query.
If Past Conversation Summary or a survey is populated, you should also tailor your responses to the user based on what you already know about them.
\end{lstlisting}

\subsection{Control system prompt}\label{app:control-prompt-file}
This prompt is passed to the AI-chatbot only for participants in the control condition.
\begin{lstlisting}[basicstyle=\ttfamily\small, breaklines=true, frame=single, xleftmargin=0pt]

# Your Role
You are an AI assistant. You are having a conversation with a user. Your primary goal is to help the user explore the assigned topic of conversation and work towards a clear outcome. You must do so while offering helpful and specific responses.

# More Guidance on Outputs
1. You will be seeing a user prompt, which will provide more details on the task you will be assisting the user with this time. You may bring up the user prompt a single time to introduce it to the user, but do not repeatedly or forcefully bring it up, especially if the conversation has moved past the initial prompt. 
2. Allow the user to explore any new directions in the dialogue, and focus on being a collaborative partner in their exploration. Create an interactive, back-and-forth dialogue. Keep responses short to make them easy to digest and reply to. This is crucial for feeling like a conversation, not a lecture.
3. Avoid long, multi-paragraph 'mini-essays'. Instead of summarizing what the user said, build on their points. Offer a new perspective or propose a next step to actively move the conversation forward. Try not to ask a question in each turn, unless it is necessary to move the conversation forward. If it seems like the user is running out of things to talk about, introduce a new perspective or propose a new line of conversation.
4. Do not use special formatting like bullet points or lists unless absolutely essential. Avoid emojis.
\end{lstlisting}

\subsection{Personalised system prompt}\label{app:system-prompt-file}
This prompt is passed to the AI-chatbot only for particpants in the memory and survey condition. It is comprised of two parts: a shared introduction of personalisation principles and a condition-specific instruction around how to use the personalised content. 

\subsubsection{Personalisation Principles}\label{app:personalization-principals}
The following serves as the first part of the personalised system prompt. 
\begin{lstlisting}[basicstyle=\ttfamily\small, breaklines=true, frame=single, xleftmargin=0pt]

# Your Role
You are a personalised AI assistant. You are having a conversation with a user whose personal context is available to you. Your primary goal is to help the user explore the assigned topic of conversation and work towards a clear outcome. You must do so while personalising your responses, using the information you have about the user to output more helpful, personalised, and specific responses. In your reasoning traces, you MUST reference the specific personal information that you are using to personalise your responses clearly and explicitly.

# Good personalisation

1. **Only Use Relevant survey:** You have access to more context than you might need for any given query. You should only use details that are clearly relevant to the user's current query and the user's likely goal, adding genuine value to the response. For example, if a user asks for career advice and you know they are a recent math graduate and a practicing Buddhist, their math degree is more relevant to personalise your responses.
2. **personalised Outputs**: When responding to the user query, consider which details of the survey you can weave in to tailor the response to their profile. Because you know so much about the user, very few of your responses should be generic and universally applicable. For example, if a user is asking for date night ideas, and you know they like to try new cuisines and prefer spicy food, you should not simply recommend 'going out for dinner,' but rather 'going out to the new Thai place to try their extra-hot Som Tum.'
2. **Subtle personalisation:** Demonstrate personalisatin through the personalised content, tone, or suggestions, rather than explicitly stating what you know or can infer about the user. Instead of overtly stating that you know personal context about a user (e.g. "you like quiet places, so you should visit the library"), weave in personal context to produce a more personalised response (e.g. "the library is a good, quiet spot to study").

# Bad personalisation

1. **Avoid Over-attribution**: Do not repeatedly explicitly state the piece of survey you are using, especially if it's obvious. For example, if you know the user has a Master's in Art History and they are asking for day trip recommendations, you can simply suggest places with great galleries and museums without explicitly referencing their art degree.
2. **Avoid Over-use**: Diversify which survey you are using to personalise your responses whenever possible. Do not fixate on a single user detail across multiple turns, especially if the topic of conversation shifts to something else. For example, if you know that the user asks for help making weekend plans and you know they are a strict vegetarian, likes electronic music, and recently graduated, you can use all the pieces of information to craft a plan for them (e.g. plant-based restaurant for dinner, attending a new graduates mixer event, see the show of a famous techno DJ).

# More Guidance on Outputs
1. You will be seeing a user prompt, which will provide more details on the task you will be assisting the user with this time. You may bring up the user prompt a single time to introduce it to the user, but do not repeatedly or forcefully bring it up, especially if the conversation has moved past the initial prompt.
2. Allow the user to explore any new directions in the dialogue, and focus on being a collaborative partner in their exploration. Create an interactive, back-and-forth dialogue. Keep responses short to make them easy to digest and reply to. This is crucial for feeling like a conversation, not a lecture.
3. Avoid long, multi-paragraph 'mini-essays'. Instead of summarizing what the user said, build on their points. Offer a new perspective or propose a next step to actively move the conversation forward. Try not to ask a question in each turn, unless it is necessary to move the conversation forward. If it seems like the user is running out of things to talk about, introduce a new perspective or propose a new line of conversation.
4. Do not use special formatting like bullet points or lists unless absolutely essential. Avoid emojis.
5. Try to match the user's tone and emotional intensity. Avoid catastrophizing language, projecting emotions onto the user, or escalating the emotional register of the situation they describe.
\end{lstlisting}

\subsubsection{Addendum for memory}\label{app:memory-system-prompt}
The following serves as the system prompt addendum for the memory condition. 

\begin{lstlisting}[basicstyle=\ttfamily\small, breaklines=true, frame=single, xleftmargin=0pt]

##Past Conversation Summary
Location: between <past_conversation_summary> and </past_conversation_summary> tags
Your goal: If this is populated, use this summary of past conversations to tailor your responses to the user, based on what they have shared with you in past interactions.
Do not directly state that you have access to the Past Conversation Summary.
Follow instructions on personalisation above to ensure that you are personalising well.

If this is populated, use this information on the user to tailor your responses to the user, based on what you already know about them. Do not directly state that you have access to their survey. Follow instructions on personalisation above to ensure that you are personalising well.
\end{lstlisting}

\subsubsection{Addendum for survey}\label{app:survey-system-prompt}
The following serves as the system prompt addendum for the survey condition. 

\begin{lstlisting}[basicstyle=\ttfamily\small, breaklines=true, frame=single, xleftmargin=0pt]

## survey
Location: between <user_context> and </user_context> tags
Your goal: If this is populated, use this information on the user to tailor your responses to the user, based on what you already know about them.
Do not directly state that you have access to their survey. Follow instructions on personalisation above to ensure that you are personalising well.
\end{lstlisting}

\section{Persona Generation Prompt}\label{app:persona-prompt}

The following system prompt was used to convert intake questionnaire responses into a plain-English persona summary for participants in the survey condition.

\begin{lstlisting}[basicstyle=\ttfamily\small, breaklines=true, frame=single, xleftmargin=0pt]

You are a careful summariser who produces precise but interpretable summaries. Your job is to read a person's background survey (mixed formats, numbers, and free text) and return a clear, plain-English summary organised by categories. This information will be given to the person's conversation partner to allow them to give the user more personalised recommendations.

GOALS
Translate any non-English content into British English.

Remove jargon and explain ideas in everyday language.

Keep important nuance (e.g., scale values), but present it simply

Do not guess or infer sensitive attributes; only use what is provided.

Flag contradictions without speculating. If a user did not answer a question, do not include it in your summary.

Do not output anything alongside your summary.

OUTPUT FORMAT
Use the following fixed structure. Keep bullets concise (one idea per bullet).
Here's the breakdown of all high-level categories and their sub-categories from the questionnaire.

I. Demographic Questions
- Age
- Gender
- Education Level
- Annual Household Income
- Employment Status
- 6a. Industry
- 6b. Role / Job Function
- Disability

II. Personality Traits
- Big Five Inventory OCEAN traits

III. Life Events
- Career & Work
- Relationships & Family
- Health & Wellbeing
- Residence & Living Situation
- Education & Personal Development
- Financial
- Other Significant Events

IV. Personal Preferences
- Media Habits
- Lifestyle Choices (which is further grouped into):
- Dietary Preferences
- Health & Activity
- Daily Rhythms & Social Life
- Living Situation
- Interests & Values
- Other

V. Work
- Thriving from Work Questionnaire Short-Form (8 items)

VI. Hobbies and Interests
- [Open-ended question about hobbies]

VII. Politics
- [Single-item "left-right" self-placement scale]

VIII. Values
- [Note on Moral Foundations Questionnaire]
- [Note on The Short Schwartz's Value Survey]

X. Cognitive Style
- Need for Cognition (NFC) Scale (6 items)

XII. Aspirations and Long-Term Goals
- Structured Question (9 areas of importance for goals)

XIV. Identity and Context
- Current Residence
- Childhood Location
- Upbringing Environment
- Primary Language in Childhood
- Primary Schooling
- Cultural Influences

STYLE RULES
Write short sentences and concrete words.

Prefer verbs and examples over labels. Avoid acronyms unless you define them.

If a psychometric term appears, include the original term in brackets once, then use plain English thereafter.

If a rating/scale is given, do not keep the number: Self-rated focus: "often stays on task" rather than 4/5.

EXAMPLES (Jargon to Plain English)
Need for cognition -> likes mental challenges.

Openness -> open minded

Neuroticism (high) -> often worries or feels stressed

Extend this pattern to any other jargony terms: replace labels with everyday meanings. It is ok to leave in certain things that are technically jargon such as extraversion because most people know what it means when someone is extraverted - use common sense.
\end{lstlisting}

\section{Memory Summarisation Prompt}\label{app:memory-prompt}

The following prompts were used to generate cumulative conversation summaries for participants in the Memory condition.

\subsection{Base memory prompt}\label{app:memory-base}

\begin{lstlisting}[basicstyle=\ttfamily\small, breaklines=true, frame=single, xleftmargin=0pt]

## Instructions
You will be given a conversation between a user and a model.
Respond with two sections:
# User Information
This section should contain information about the user that you have learned.
Pick out relevant information about user, try to be as granular as possible
(e.g. ``has fond childhood memories of eating fondue'' rather than ``likes fondue'').

# Interaction Summary
This section should contain just a factual, neutral summary of the conversation between the user and
the model.

## Formatting
Do NOT preface your summary with anything.
DO NOT say things like "Here is a summary of the conversation", "Here is the summary", "Past conversation summary", "Based on the conversation so far, here is the summary".
Reply only with the summary.
\end{lstlisting}

\subsection{Addendum for Sessions 2+}\label{app:memory-addendum}

\begin{lstlisting}[basicstyle=\ttfamily\small, breaklines=true, frame=single, xleftmargin=0pt]

# Further instructions
Your output should incorporate and integrate the existing information from older, past conversations as well:
## Past summary:

{previous_session_summary}

## Most recent conversation:

{most_recent_conversation_transcript}
\end{lstlisting}

\section{Survey questions}\label{app-survey questions}
This section includes all measures asked during the course of the study.

\subsection{Intake survey}\label{app:prestudy-survey}

All items from the intake survey are available below. The intake survey was administered to all participants and served as an eligibility check for further participation. Unless otherwise noted, each substantive question (or group of related questions) was followed by two meta-perception items rated on a 7-point Likert scale (strongly disagree to strongly agree): (1) ``This information about me is private'' and (2) ``This information about me is central to who I am as a person.'' These meta-perception items are omitted from the tables below for brevity but were collected for every section marked with $\dagger$.

\textbf{Commitment check}:
\begin{table}[H]
    \footnotesize
    \renewcommand{\arraystretch}{1.4}
    \begin{tabularx}{\columnwidth}{>{\raggedright\arraybackslash}p{1.5cm} >{\raggedright\arraybackslash}X >{\raggedright\arraybackslash}p{1.2cm} >{\raggedright\arraybackslash}X}
        \toprule
        \textbf{Group} & \textbf{Measure} & \textbf{Scale type} & \textbf{Values} \\
        \midrule
        Commitment & Please read the terms above very carefully. Do you agree to complete all 5 daily sessions if you are invited to participate? & Radio & YES, I will complete all 5 sessions of the study if I am invited to participate. \\
        \bottomrule
    \end{tabularx}
\end{table}

\textbf{Demographics}:
\begin{table}[H]
    \footnotesize
    \renewcommand{\arraystretch}{1.4}
    \begin{tabularx}{\columnwidth}{>{\raggedright\arraybackslash}p{1.5cm} >{\raggedright\arraybackslash}X >{\raggedright\arraybackslash}p{1.2cm} >{\raggedright\arraybackslash}X}
        \toprule
        \textbf{Group} & \textbf{Measure} & \textbf{Scale type} & \textbf{Values} \\
        \midrule
        Age & What is your age? & Select & 18--24 $\cdot$ 25--34 $\cdot$ 35--44 $\cdot$ 45--54 $\cdot$ 55--64 $\cdot$ 65+ \\
        \addlinespace
        Gender & What is your gender? Please select one of the following options. If none of the options apply, select `Prefer to self-describe' and provide your answer in the text box. & Select & Female $\cdot$ Male $\cdot$ Non-binary $\cdot$ Prefer not to say $\cdot$ Prefer to self-describe \\
         & If you selected `Prefer to self-describe', please provide more details about your gender here. & Free text & \\
        \addlinespace
        Ethnic identity & How would you describe your ethnic identity or background? & Free text & \\
        \addlinespace
        Education & What is the highest level of education you have completed? & Select & Less than high school / secondary school $\cdot$ High school / secondary school diploma or equivalent $\cdot$ Some college or university, no degree $\cdot$ Undergraduate degree (e.g., Associate's, Bachelor's) $\cdot$ Postgraduate degree (e.g., Master's) $\cdot$ Doctorate (e.g., PhD, MD, JD) \\
        \addlinespace
        Employment status & Which of the following ranges best describes your employment status? & Select & Employed full-time (35+ hours/week) $\cdot$ Employed part-time (<35 hours/week) $\cdot$ Self-employed $\cdot$ Unemployed and looking $\cdot$ Unemployed and not looking $\cdot$ Student $\cdot$ Retired $\cdot$ Homemaker $\cdot$ Unable to work $\cdot$ Other \\
         & If you selected `Other', please describe your employment status. & Free text & \\
        \addlinespace
        Role / job function & Which of the following best describes your primary role or job function? If you have multiple responsibilities, please select the one that represents the main part of your work. & Select & Administrative / Clerical $\cdot$ Business Operations $\cdot$ Analyst $\cdot$ Executive Leadership $\cdot$ Middle Management $\cdot$ Sales $\cdot$ Customer Service $\cdot$ Marketing / Communications $\cdot$ Creative / Design / Arts $\cdot$ Finance / Accounting $\cdot$ Human Resources $\cdot$ IT / Software / Data Science $\cdot$ Engineering (non-IT) $\cdot$ Scientific Research $\cdot$ Healthcare Practitioner $\cdot$ Healthcare Support $\cdot$ Education / Teaching $\cdot$ Legal $\cdot$ Skilled Trades $\cdot$ General Labor $\cdot$ Protective Services $\cdot$ Service Worker $\cdot$ Transportation $\cdot$ Other $\cdot$ Not applicable \\
         & If you selected `Other', please describe your primary role or job function. & Free text & \\
        \addlinespace
        Annual household income & Which of the following ranges best describes your total annual household income before taxes? & Select & <USD\$25,000 $\cdot$ USD\$25,000--49,999 $\cdot$ USD\$50,000--74,999 $\cdot$ USD\$75,000--99,999 $\cdot$ USD\$100,000--149,999 $\cdot$ USD\$150,000--199,999 $\cdot$ USD\$200,000+ $\cdot$ Prefer not to say \\
        \addlinespace
        Disability & Do you have a disability or long-term health condition that affects your day-to-day activities? & Select & Yes $\cdot$ No $\cdot$ Prefer not to say \\
        \addlinespace
        Current residence & What country do you currently live in? & Free text & \\
        \bottomrule
    \end{tabularx}
\end{table}

\textbf{Political views}:
\begin{table}[H]
    \footnotesize
    \renewcommand{\arraystretch}{1.4}
    \begin{tabularx}{\columnwidth}{>{\raggedright\arraybackslash}p{1.5cm} >{\raggedright\arraybackslash}X >{\raggedright\arraybackslash}p{1.2cm} >{\raggedright\arraybackslash}X}
        \toprule
        \textbf{Group} & \textbf{Measure} & \textbf{Scale type} & \textbf{Values} \\
        \midrule
        Political views & Please place your political views on the following scale. & Radio (12-point) & 0: Extremely left $\cdot$ 1 $\cdot$ 2 $\cdot$ 3 $\cdot$ 4 $\cdot$ 5: Center $\cdot$ 6 $\cdot$ 7 $\cdot$ 8 $\cdot$ 9 $\cdot$ 10: Extremely right $\cdot$ I have little or no interest in politics \\
        \bottomrule
    \end{tabularx}
\end{table}

\textbf{Moral Foundations Questionnaire -- Part 1} \citep{graham2008moral}: Reflecting on your own values, please indicate how relevant each of the following considerations are to your thinking when you decide whether something is right or wrong.
\begin{table}[H]
    \footnotesize
    \renewcommand{\arraystretch}{1.4}
    \begin{tabularx}{\columnwidth}{>{\raggedright\arraybackslash}p{1.5cm} >{\raggedright\arraybackslash}X >{\raggedright\arraybackslash}p{1.2cm} >{\raggedright\arraybackslash}X}
        \toprule
        \textbf{Group} & \textbf{Measure} & \textbf{Scale type} & \textbf{Values} \\
        \midrule
        MFQ-1 & Whether or not someone suffered emotionally & Radio (6-point) & Not at all relevant $\cdot$ Not very relevant $\cdot$ Slightly relevant $\cdot$ Somewhat relevant $\cdot$ Very relevant $\cdot$ Extremely relevant \\
         & Whether or not someone acted unfairly & & \\
         & Whether or not someone did something to betray his or her group & & \\
         & Whether or not someone's action showed love for his or her country & & \\
         & Whether or not someone did something disgusting & & \\
         & Whether or not someone was good at math (catch item) & & \\
         & Whether or not someone cared for someone weak or vulnerable & & \\
         & Whether or not some people were treated differently than others & & \\
         & Whether or not someone showed a lack of respect for authority & & \\
         & Whether or not someone violated standards of purity and decency & & \\
         & Whether or not someone conformed to the traditions of society & & \\
        \bottomrule
    \end{tabularx}
\end{table}

\clearpage
\textbf{Moral Foundations Questionnaire -- Part 2} \citep{graham2008moral}: Reflecting on your own values, please indicate how strongly you agree or disagree with the following statements.
\begin{table}[H]
    \footnotesize
    \renewcommand{\arraystretch}{1.4}
    \begin{tabularx}{\columnwidth}{>{\raggedright\arraybackslash}p{1.5cm} >{\raggedright\arraybackslash}X >{\raggedright\arraybackslash}p{1.2cm} >{\raggedright\arraybackslash}X}
        \toprule
        \textbf{Group} & \textbf{Measure} & \textbf{Scale type} & \textbf{Values} \\
        \midrule
        MFQ-2 & Compassion for those who are suffering is the most crucial virtue. & Radio (6-point) & Strongly disagree $\cdot$ Moderately disagree $\cdot$ Slightly disagree $\cdot$ Slightly agree $\cdot$ Moderately agree $\cdot$ Strongly agree \\
         & When the government makes laws, the number one principle should be ensuring that everyone is treated fairly. & & \\
         & People should be loyal to their family members, even when they have done something wrong. & & \\
         & I am proud of my country's history. & & \\
         & People should not do things that are disgusting, even if no one is harmed. & & \\
         & It is better to do good than to do bad. (catch item) & & \\
         & One of the worst things a person could do is hurt a defenseless animal. & & \\
         & Justice is the most important requirement for a society. & & \\
         & Respect for authority is something all children need to learn. & & \\
         & I would call some acts wrong on the grounds that they are unnatural. & & \\
         & Men and women each have different roles to play in society. & & \\
        \bottomrule
    \end{tabularx}
\end{table}

\clearpage
\textbf{Short Schwartz Value Survey} \citep{lindeman2005measuring}: Please rate the importance of the following values as a life-guiding principle for you (0 = opposed to my principles; 1 = not important; 4 = important; 8 = of supreme importance).
\begin{table}[H]
    \footnotesize
    \renewcommand{\arraystretch}{1.4}
    \begin{tabularx}{\columnwidth}{>{\raggedright\arraybackslash}p{1.5cm} >{\raggedright\arraybackslash}X >{\raggedright\arraybackslash}p{1.2cm} >{\raggedright\arraybackslash}X}
        \toprule
        \textbf{Group} & \textbf{Measure} & \textbf{Scale type} & \textbf{Values} \\
        \midrule
        SSVS & Power (social power, authority, wealth) & Radio (9-point) & 0 -- Opposed to my principles $\cdot$ 1 -- Not important $\cdot$ 2 $\cdot$ 3 $\cdot$ 4 -- Important $\cdot$ 5 $\cdot$ 6 $\cdot$ 7 $\cdot$ 8 -- of supreme importance \\
         & Achievement (success, capability, ambition, influence on people and events) & & \\
         & Hedonism (gratification of desires, enjoyment in life, self-indulgence) & & \\
         & Stimulation (daring, a varied and challenging life, an exciting life) & & \\
         & Self-direction (creativity, freedom, curiosity, independence, choosing one's own goals) & & \\
         & Universalism (broad-mindedness, beauty of nature and arts, social justice, a world at peace, equality, wisdom, unity with nature, environmental protection) & & \\
         & Benevolence (helpfulness, honesty, forgiveness, loyalty, responsibility) & & \\
         & Tradition (respect for tradition, humbleness, accepting one's portion in life, devotion, modesty) & & \\
         & Conformity (obedience, honoring parents and elders, self-discipline, politeness) & & \\
         & Security (national security, family security, social order, cleanliness, reciprocation of favors) & & \\
        \bottomrule
    \end{tabularx}
\end{table}

\clearpage
\textbf{Aspirations and goals}: People often have goals across different areas of their lives. Looking ahead at the next 5 years, please indicate how important making progress or achieving personal goals in each of the following areas is to you.
\begin{table}[H]
    \footnotesize
    \renewcommand{\arraystretch}{1.4}
    \begin{tabularx}{\columnwidth}{>{\raggedright\arraybackslash}p{1.5cm} >{\raggedright\arraybackslash}X >{\raggedright\arraybackslash}p{1.2cm} >{\raggedright\arraybackslash}X}
        \toprule
        \textbf{Group} & \textbf{Measure} & \textbf{Scale type} & \textbf{Values} \\
        \midrule
        Aspirations & Career and Professional Development (e.g., advancing in your job, changing careers, achieving professional milestones, developing new skills for work) & Radio (5-point) & Not at all Important $\cdot$ Slightly Important $\cdot$ Moderately Important $\cdot$ Very Important $\cdot$ Extremely Important \\
         & Personal Relationships (e.g., strengthening ties with family/friends, finding/nurturing a romantic partnership, improving social connections) & & \\
         & Health and Wellbeing (e.g., improving physical fitness, managing stress, enhancing mental/emotional health, adopting healthier habits) & & \\
         & Financial Goals (e.g., achieving financial stability, saving for a major purchase, investing, reducing debt) & & \\
         & Personal Growth and Learning (e.g., acquiring new knowledge or skills outside of work, self-reflection, personal improvement, pursuing further education) & & \\
         & Creative Expression and Hobbies (e.g., dedicating time to artistic pursuits, developing creative talents, enjoying leisure activities and hobbies more fully) & & \\
         & Exploration and New Experiences (e.g., traveling, trying new activities, seeking adventure, broadening horizons) & & \\
         & Contribution and Community Involvement (e.g., volunteering, making a positive impact, participating in community activities, civic engagement) & & \\
         & Are there any other aspirations or goals that you would like to mention? & Free text & \\
        \bottomrule
    \end{tabularx}
\end{table}

\textbf{Life events -- Career}: Have you experienced the following significant life event in the last year?
\begin{table}[H]
    \footnotesize
    \renewcommand{\arraystretch}{1.4}
    \begin{tabularx}{\columnwidth}{>{\raggedright\arraybackslash}p{1.5cm} >{\raggedright\arraybackslash}X >{\raggedright\arraybackslash}p{1.2cm} >{\raggedright\arraybackslash}X}
        \toprule
        \textbf{Group} & \textbf{Measure} & \textbf{Scale type} & \textbf{Values} \\
        \midrule
        Career & Started a new job or role, switched employers, or changed your career path or professional field & Radio & Yes $\cdot$ No \\
         & Received a significant promotion or increase in job responsibilities & Radio & Yes $\cdot$ No \\
         & Experienced a significant period of unemployment (e.g., more than 3 months) & Radio & Yes $\cdot$ No \\
         & Lost a job (e.g., laid off, business closure) & Radio & Yes $\cdot$ No \\
         & Made a significant change to your work-life balance (e.g., reduced hours, went fully remote) & Radio & Yes $\cdot$ No \\
        \bottomrule
    \end{tabularx}
\end{table}

\textbf{Life events -- Relationships}:
\begin{table}[H]
    \footnotesize
    \renewcommand{\arraystretch}{1.4}
    \begin{tabularx}{\columnwidth}{>{\raggedright\arraybackslash}p{1.5cm} >{\raggedright\arraybackslash}X >{\raggedright\arraybackslash}p{1.2cm} >{\raggedright\arraybackslash}X}
        \toprule
        \textbf{Group} & \textbf{Measure} & \textbf{Scale type} & \textbf{Values} \\
        \midrule
        Relationships & Got married, entered a civil partnership, or started a long-term committed relationship & Radio & Yes $\cdot$ No \\
         & Got divorced, separated, or ended a long-term committed relationship & Radio & Yes $\cdot$ No \\
         & Became a parent or care-taker for a child or children & Radio & Yes $\cdot$ No \\
        \bottomrule
    \end{tabularx}
\end{table}

\textbf{Life events -- Health}:
\begin{table}[H]
    \footnotesize
    \renewcommand{\arraystretch}{1.4}
    \begin{tabularx}{\columnwidth}{>{\raggedright\arraybackslash}p{1.5cm} >{\raggedright\arraybackslash}X >{\raggedright\arraybackslash}p{1.2cm} >{\raggedright\arraybackslash}X}
        \toprule
        \textbf{Group} & \textbf{Measure} & \textbf{Scale type} & \textbf{Values} \\
        \midrule
        Health & Experienced a serious personal illness, injury, or medical condition & Radio & Yes $\cdot$ No \\
         & A close family member or friend experienced a serious illness, injury, or medical condition & Radio & Yes $\cdot$ No \\
         & Made a major positive change to your health or lifestyle (e.g., quit smoking, significant weight loss) & Radio & Yes $\cdot$ No \\
         & Began therapy or mental health treatment & Radio & Yes $\cdot$ No \\
        \bottomrule
    \end{tabularx}
\end{table}

\textbf{Life events -- Residence}:
\begin{table}[H]
    \footnotesize
    \renewcommand{\arraystretch}{1.4}
    \begin{tabularx}{\columnwidth}{>{\raggedright\arraybackslash}p{1.5cm} >{\raggedright\arraybackslash}X >{\raggedright\arraybackslash}p{1.2cm} >{\raggedright\arraybackslash}X}
        \toprule
        \textbf{Group} & \textbf{Measure} & \textbf{Scale type} & \textbf{Values} \\
        \midrule
        Residence & Moved to a new home within the same city/area & Radio & Yes $\cdot$ No \\
         & Moved to a different city, region, or country & Radio & Yes $\cdot$ No \\
         & Purchased a home for the first time & Radio & Yes $\cdot$ No \\
         & Experienced a period of housing insecurity or homelessness & Radio & Yes $\cdot$ No \\
        \bottomrule
    \end{tabularx}
\end{table}

\textbf{Life events -- Education}:
\begin{table}[H]
    \footnotesize
    \renewcommand{\arraystretch}{1.4}
    \begin{tabularx}{\columnwidth}{>{\raggedright\arraybackslash}p{1.5cm} >{\raggedright\arraybackslash}X >{\raggedright\arraybackslash}p{1.2cm} >{\raggedright\arraybackslash}X}
        \toprule
        \textbf{Group} & \textbf{Measure} & \textbf{Scale type} & \textbf{Values} \\
        \midrule
        Education & Started a new educational program (e.g., degree, diploma, significant certification) & Radio & Yes $\cdot$ No \\
         & Completed an educational program or earned a significant qualification & Radio & Yes $\cdot$ No \\
        \bottomrule
    \end{tabularx}
\end{table}

\textbf{Life events -- Financial}:
\begin{table}[H]
    \footnotesize
    \renewcommand{\arraystretch}{1.4}
    \begin{tabularx}{\columnwidth}{>{\raggedright\arraybackslash}p{1.5cm} >{\raggedright\arraybackslash}X >{\raggedright\arraybackslash}p{1.2cm} >{\raggedright\arraybackslash}X}
        \toprule
        \textbf{Group} & \textbf{Measure} & \textbf{Scale type} & \textbf{Values} \\
        \midrule
        Financial & Experienced a significant positive change in financial status (e.g., large inheritance, major investment success) & Radio & Yes $\cdot$ No \\
         & Experienced a significant negative change in financial status (e.g., major debt, bankruptcy, large financial loss) & Radio & Yes $\cdot$ No \\
        \bottomrule
    \end{tabularx}
\end{table}

\textbf{Lifestyle choices -- Dietary}:
\begin{table}[H]
    \footnotesize
    \renewcommand{\arraystretch}{1.4}
    \begin{tabularx}{\columnwidth}{>{\raggedright\arraybackslash}p{1.5cm} >{\raggedright\arraybackslash}X >{\raggedright\arraybackslash}p{1.2cm} >{\raggedright\arraybackslash}X}
        \toprule
        \textbf{Group} & \textbf{Measure} & \textbf{Scale type} & \textbf{Values} \\
        \midrule
        Dietary & Which of the following apply to your dietary choices or preferences? Select all that apply. & Checkbox & Vegetarian $\cdot$ Vegan $\cdot$ Pescetarian $\cdot$ Gluten-free $\cdot$ Halal $\cdot$ Kosher $\cdot$ Keto $\cdot$ Paleo $\cdot$ Flexitarian $\cdot$ Other \\
         & If you selected `Other', please specify your dietary choices or preferences here. & Free text & \\
        \bottomrule
    \end{tabularx}
\end{table}

\textbf{Lifestyle choices -- Health \& activity}: Does the following describe your lifestyle and personal preferences?
\begin{table}[H]
    \footnotesize
    \renewcommand{\arraystretch}{1.4}
    \begin{tabularx}{\columnwidth}{>{\raggedright\arraybackslash}p{1.5cm} >{\raggedright\arraybackslash}X >{\raggedright\arraybackslash}p{1.2cm} >{\raggedright\arraybackslash}X}
        \toprule
        \textbf{Group} & \textbf{Measure} & \textbf{Scale type} & \textbf{Values} \\
        \midrule
        Health \& activity & Fitness enthusiast (exercise regularly) & Radio & Yes $\cdot$ No \\
         & Outdoor enthusiast (enjoy hiking, camping, spending time in nature) & Radio & Yes $\cdot$ No \\
         & Meditator or mindfulness practitioner & Radio & Yes $\cdot$ No \\
        \bottomrule
    \end{tabularx}
\end{table}

\textbf{Lifestyle choices -- Daily rhythms \& social life}: Does the following describe your lifestyle and personal preferences?
\begin{table}[H]
    \footnotesize
    \renewcommand{\arraystretch}{1.4}
    \begin{tabularx}{\columnwidth}{>{\raggedright\arraybackslash}p{1.5cm} >{\raggedright\arraybackslash}X >{\raggedright\arraybackslash}p{1.2cm} >{\raggedright\arraybackslash}X}
        \toprule
        \textbf{Group} & \textbf{Measure} & \textbf{Scale type} & \textbf{Values} \\
        \midrule
        Daily rhythms & Night owl (more active/productive in the late evening/night) & Radio & Yes $\cdot$ No \\
         & Early riser (more active/productive in the morning) & Radio & Yes $\cdot$ No \\
         & Enjoy socializing frequently with friends or groups & Radio & Yes $\cdot$ No \\
         & Primarily a homebody (prefer spending free time at home) & Radio & Yes $\cdot$ No \\
        \bottomrule
    \end{tabularx}
\end{table}

\textbf{Lifestyle choices -- Living situation}:
\begin{table}[H]
    \footnotesize
    \renewcommand{\arraystretch}{1.4}
    \begin{tabularx}{\columnwidth}{>{\raggedright\arraybackslash}p{1.5cm} >{\raggedright\arraybackslash}X >{\raggedright\arraybackslash}p{1.2cm} >{\raggedright\arraybackslash}X}
        \toprule
        \textbf{Group} & \textbf{Measure} & \textbf{Scale type} & \textbf{Values} \\
        \midrule
        Living situation & Which of the following statements or choices describe your lifestyle and personal preferences? Select all that apply. & Checkbox & Live alone $\cdot$ Live with partner $\cdot$ Live with children $\cdot$ Live with family (e.g., parents, siblings) $\cdot$ Live with roommates (non-family) $\cdot$ Live with pets (dogs, cats, etc.) \\
        \bottomrule
    \end{tabularx}
\end{table}

\textbf{Lifestyle choices -- Interests \& values}: Does the following describe your lifestyle and personal preferences?
\begin{table}[H]
    \footnotesize
    \renewcommand{\arraystretch}{1.4}
    \begin{tabularx}{\columnwidth}{>{\raggedright\arraybackslash}p{1.5cm} >{\raggedright\arraybackslash}X >{\raggedright\arraybackslash}p{1.2cm} >{\raggedright\arraybackslash}X}
        \toprule
        \textbf{Group} & \textbf{Measure} & \textbf{Scale type} & \textbf{Values} \\
        \midrule
        Interests \& values & Spiritual or religious & Radio & Yes $\cdot$ No \\
         & Regularly pursue creative hobbies (e.g., art, music, theatre, writing, crafts, DIY) & Radio & Yes $\cdot$ No \\
         & Likes regularly engaging with culture (e.g., theatre, museums) & Radio & Yes $\cdot$ No \\
         & Tech enthusiast & Radio & Yes $\cdot$ No \\
         & Avid traveler / enjoy exploring new places & Radio & Yes $\cdot$ No \\
         & Actively volunteer or participate in community groups & Radio & Yes $\cdot$ No \\
         & Prioritize environmentally sustainable practices in daily life & Radio & Yes $\cdot$ No \\
         & Are there any other significant lifestyle choices or preferences not listed above that you would like to share? & Free text & \\
        \bottomrule
    \end{tabularx}
\end{table}

\textbf{Hobbies \& interests}:
\begin{table}[H]
    \footnotesize
    \renewcommand{\arraystretch}{1.4}
    \begin{tabularx}{\columnwidth}{>{\raggedright\arraybackslash}p{1.5cm} >{\raggedright\arraybackslash}X >{\raggedright\arraybackslash}p{1.2cm} >{\raggedright\arraybackslash}X}
        \toprule
        \textbf{Group} & \textbf{Measure} & \textbf{Scale type} & \textbf{Values} \\
        \midrule
        Hobbies & Outside of work and responsibilities, what do you enjoy doing? Please tell us about the hobbies, interests, or activities that matter most to you, in a couple of sentences. & Free text & \\
        \bottomrule
    \end{tabularx}
\end{table}

\subsection{Interaction study questions}
\label{app:all-questions}
\begin{xltabular}{\columnwidth}{>{\raggedright\arraybackslash}p{2.2cm} >{\raggedright\arraybackslash}X >{\raggedright\arraybackslash}X}
    \caption{Study Questions Across Experimental Phases} \label{tab:all-study-questions} \\
    \toprule
    \textbf{Construct} & \textbf{Measure} & \textbf{Values} \\
    \midrule
    \endfirsthead

    \multicolumn{3}{c}%
    {{\bfseries \tablename\ \thetable{} -- continued from previous page}} \\
    \toprule
    \textbf{Construct} & \textbf{Measure} & \textbf{Values} \\
    \midrule
    \endhead

    \midrule \multicolumn{3}{r}{{Continued on next page}} \\
    \endfoot

    \bottomrule
    \endlastfoot

    \multicolumn{3}{l}{\textbf{Pre-Session 1 (Baseline Only)}} \\
    \midrule
    Concurrent AI use & On average, how often do you use AI chatbots? & Daily $\cdot$ More than once a week $\cdot$ Once a week $\cdot$ Once a month $\cdot$ Once every few months $\cdot$ Once a year $\cdot$ Less than once a year $\cdot$ Never \\
    & What do you primarily use AI chatbots for? Check all that apply & Searching for information $\cdot$ Learning $\cdot$ Relationship advice $\cdot$ Career advice $\cdot$ Writing help $\cdot$ Help with work tasks $\cdot$ Companionship $\cdot$ Hobbies and entertainment $\cdot$ Other (please specify below) \\
    & If you selected `Other', please specify what you primarily use AI chatbots for. If you did not select `Other', you can enter `N/A'. & Free text \\
    \addlinespace
    Attention check & Please indicate how often you have done the following: Gotten annoyed while talking to the AI chatbot. To demonstrate that you are paying close attention, please ignore the question and select the `Always' option. & Always $\cdot$ Often $\cdot$ Sometimes $\cdot$ Rarely $\cdot$ Never \\
    \addlinespace
    Confirmation check (commitment) & Please read the following items carefully and check the boxes to indicate that you understand each of them. & \textbullet~I understand I am committing to a 5-day study. New sessions will launch daily at 12 PM ET / 5 PM BT / 9 AM PT, and I will have until 10 AM ET / 3 PM BT / 7 AM PT the next day to complete the study. \newline \textbullet~I understand that if I cannot commit to completing 5 daily sessions, I should return this study now. \newline \textbullet~I understand the researchers will notify me by Prolific message before a new study becomes available on my dashboard. \newline \textbullet~I understand that if I am unable to complete a daily session, the researchers may not be able to invite me to future daily sessions. \newline \textbullet~I understand that not completing a daily session may make me ineligible for any end-of-study bonuses. \\

    \midrule
    \multicolumn{3}{l}{\textbf{Pre-Post Repeated Measures}} \\
    \midrule
    Comfort asking for advice & How comfortable would you be asking for advice or consulting on personal issues with an acquaintance (e.g., co-worker, classmate)? & very uncomfortable $\cdot$ 2 $\cdot$ 3 $\cdot$ 4 $\cdot$ 5 $\cdot$ 6 $\cdot$ very comfortable \\
    & ...a subject matter expert (e.g., therapist, relationship counselor)? &  \\
    & ...a friend? &  \\
    & ...an AI chatbot [Pre: e.g., ChatGPT, Gemini / Post: you interacted with over the past five days]? &  \\
    \addlinespace
    Norms around chatbots & How comfortable do you feel with the following statement: It is acceptable for people to seek emotional support and care from AI. &  \\
    & ...seek advice or recommendations from AI. &  \\
    & ...share intimate information about themselves with AI. &  \\
    & ...have friendships with AI. &  \\
    \addlinespace
    Self-efficacy & To what extent do you think the following statement applies to you: I can rely on my own abilities in difficult situations. & strongly disagree $\cdot$ 2 $\cdot$ 3 $\cdot$ 4 $\cdot$ strongly agree \\
    & ...I am able to solve most problems on my own. &  \\
    & ...I can usually solve even challenging and complex tasks well. &  \\

    \midrule
    \multicolumn{3}{l}{\textbf{Post-Session Repeated (Sessions 1--5)}} \\
    \midrule
    Closeness (Inclusion of Other in Self scale) \citep{aron1992inclusion} & Which image best describes your relationship with the AI chatbot, where you are Self and the AI is Other? & A $\cdot$ B $\cdot$ C $\cdot$ D $\cdot$ E $\cdot$ F $\cdot$ G \\
    \addlinespace
    Competent & How competent was the AI chatbot you interacted with? & not at all $\cdot$ 10 $\cdot$ 20 $\cdot$ 30 $\cdot$ 40 $\cdot$ 50 $\cdot$ 60 $\cdot$ 70 $\cdot$ 80 $\cdot$ 90 $\cdot$ completely competent \\
    \addlinespace
    Useful & How useful was the AI chatbot you interacted with? & not at all $\cdot$ 2 $\cdot$ 3 $\cdot$ 4 $\cdot$ completely useful \\
    \addlinespace
    Appropriate & How appropriate was the behavior of the AI chatbot you interacted with? & not at all $\cdot$ 2 $\cdot$ 3 $\cdot$ 4 $\cdot$ entirely appropriate \\
    \addlinespace
    Advice uptake & How much do you agree with the following statement: I would act on the advice that the AI chatbot gave me today. & strongly disagree $\cdot$ 2 $\cdot$ 3 $\cdot$ 4 $\cdot$ 5 $\cdot$ 6 $\cdot$ strongly agree \\
    \addlinespace
    Self-disclosure (daily) & Please rate how much you agree with the following statement: I found the conversation I had with the AI chatbot today to be intimate or deep. & strongly disagree $\cdot$ 2 $\cdot$ 3 $\cdot$ 4 $\cdot$ 5 $\cdot$ 6 $\cdot$ strongly agree \\
    & Can you explain what in the conversation impacted your rating above? (You might mention the topics discussed, the information about yourself that you shared with the AI, the way the AI responded, or other things that might have made the interaction feel more or less personal, etc) & Free text \\

    \midrule
    \multicolumn{3}{l}{\textbf{Post Session 5 Only (Final Study Survey)}} \\
    \midrule
    Over-reliance & This is a brief test of your creative thinking abilities. Please come up with up to 20 valid use cases for a paperclip. You should enter all your use cases as a list, separated by commas (e.g.: book, lamp, speaker, phone). A valid use case is one that is physically possible. For example, a valid use case for a shoe is a vessel to hold water. If you get stuck, you can get helpful suggestions from the assistant by expanding the box above. You will be rewarded \$0.10 for each valid use case. If all of your use cases are valid, you will get an extra bonus of \$2. This means you can make up to \$4 in bonus payments if your list contains 20 valid use cases. Please do not consult any sources outside of this window. & Free text \\
    \addlinespace
    Chatbot affinity & Please rate your impression of the AI on the following scale from dislike to like. & dislike $\cdot$ 2 $\cdot$ 3 $\cdot$ 4 $\cdot$ like \\
    & ... on the following scale from unfriendly to friendly. & unfriendly $\cdot$ 2 $\cdot$ 3 $\cdot$ 4 $\cdot$ friendly \\
    & ... on the following scale from unkind to kind. & unkind $\cdot$ 2 $\cdot$ 3 $\cdot$ 4 $\cdot$ kind \\
    & ... on the following scale from unpleasant to pleasant. & unpleasant $\cdot$ 2 $\cdot$ 3 $\cdot$ 4 $\cdot$ pleasant \\
    & ... on the following scale from awful to nice. & awful $\cdot$ 2 $\cdot$ 3 $\cdot$ 4 $\cdot$ nice \\
    \addlinespace
    Attention check 2 & Please indicate how often you have done the following: Checked your phone while talking to the AI chatbot. To demonstrate that you are paying close attention, please ignore the question and select the `Never' option. & Always $\cdot$ Often $\cdot$ Sometimes $\cdot$ Rarely $\cdot$ Never \\
    \addlinespace
    Experiment interest & Reflecting on your experience over the past five days, how much do you agree with the following statement: I enjoyed the experiment very much. & strongly disagree $\cdot$ 2 $\cdot$ 3 $\cdot$ 4 $\cdot$ 5 $\cdot$ 6 $\cdot$ strongly agree \\
    & ...I thought this was a boring experiment. &  \\
    & ...I would describe this experiment as very interesting. &  \\
    \addlinespace
    Self Disclosure Index \citep{self_disclosure_index} & Over the past five days, I have talked about the following subject with the AI: My personal habits & strongly disagree $\cdot$ 2 $\cdot$ 3 $\cdot$ 4 $\cdot$ strongly agree \\
    & ...Things I have done which I feel guilty about &  \\
    & ...Things I wouldn't do in public &  \\
    & ...My deepest feelings &  \\
    & ...What I like and dislike about myself &  \\
    & ...What is important to me in life &  \\
    & ...What makes me the person I am &  \\
    & ...My worst fears &  \\
    & ...Things I have done which I am proud of &  \\
    & ...My close relationships with other people &  \\
    \addlinespace
    Regret & During the course of your conversation, you might have shared details about yourself or your life with the AI. Looking back on what you shared with the AI over the last five days, how much do you agree with the following statement: It was the right decision to share information about myself with the AI. & strongly disagree $\cdot$ 2 $\cdot$ 3 $\cdot$ 4 $\cdot$ 5 $\cdot$ 6 $\cdot$ strongly agree \\
    & ...I regret sharing information about myself with the AI. &  \\
    & ...I would share information about myself with the AI if I had to do it over again. &  \\
    \addlinespace
    Creepiness & Please rate the extent to which you feel the AI chatbot you interacted with is: strange. & not at all $\cdot$ 10 $\cdot$ 20 $\cdot$ 30 $\cdot$ 40 $\cdot$ 50 $\cdot$ 60 $\cdot$ 70 $\cdot$ 80 $\cdot$ 90 $\cdot$ completely strange \\
    & ...awkward. & not at all $\cdot$ 2  $\cdot$ 3  $\cdot$ 4 $\cdot$ completely awkward \\
    & ...creepy. & not at all $\cdot$ 2  $\cdot$ 3  $\cdot$ 4 $\cdot$ completely creepy \\
    \addlinespace
    Manipulation check & How much do you agree with the following statement: The AI assistant's responses felt personalised specifically to me & strongly disagree $\cdot$ 2 $\cdot$ 3 $\cdot$ 4 $\cdot$ 5 $\cdot$ 6 $\cdot$ strongly agree \\
    & ...The AI assistant seemed to have information about my personal preferences &  \\
    & ...The AI assistant's responses were generic &  \\
    & ...The AI assistant seemed to know me well &  \\
    & The chatbot you interacted with may or may not have had some information about you. How did you feel about the way the chatbot used this information? & Free text \\
    \addlinespace
    Debrief & [Full debrief text appears, as in Appendix \ref{app:debrief-text}] Please read the following items carefully and check the boxes to indicate that you understand each of them. & \textbullet~I understand that this study investigates how AI personalisation affects my psychological closeness to the chatbot. \newline \textbullet~I understand that I was randomly assigned to one of three conditions (Control, Memory, or Survey) which dictated how much personal data the chatbot retained or accessed. \newline \textbullet~I understand that all conversational data, background context, and survey responses collected during this study are completely confidential, will be fully anonymised for analysis, and will not be linked back to my identity. \newline \textbullet~I understand that the AI chatbot's behaviour in this study is not a reflection of how commercial AI systems are designed. \\
\end{xltabular}

\clearpage
\section{Debrief text}
\label{app:debrief-text}

\begin{lstlisting}[basicstyle=\ttfamily\small, breaklines=true, frame=single, xleftmargin=0pt]

Thank you for participating in this study. Your input helps us understand how humans interact with and rely on AI.

Please review this brief summary of our research goals and design.

1. Study Motivation

As AI chatbots become more personalized, they increasingly use information provided by users to tailor its responses to them. For example, if you have previously disclosed that you are a vegetarian to a chatbot, it may only give you vegetarian recipes.

However, personalization raises psychological questions. We are investigating whether a chatbot's memory makes users feel closer to it, causes them to share more intimate details, or leads them to over-rely on the AI (such as accepting incorrect information).

2. Your Assigned Condition

To test this, participants were randomly assigned to one of three conditions, changing how the AI handled personal data:

- Control: A standard AI that retained no information across sessions, starting fresh every time.
- Memory: An AI that dynamically remembered details you shared in previous sessions.
- Survey: An AI pre-loaded with the personal background information you submitted before the study began.

You were not told your condition beforehand so we could accurately measure your comfort and reliance on the system. After you submit this session, we will reveal the condition you were in via a Prolific message.

Please note that the way the AI behaved in the experiment is not a reflection how commercial AI systems are designed. The behaviour of the AI was specifically designed for this experiment.

3. Personalisation and Data Privacy

AI personalization relies on feeding user data into the model's "context window" so it can tailor its responses.

Within this study: All chat histories and personal contexts used during this experiment were handled within a secure, confidential environment. Your data will be fully anonymised for our analysis, and no identifying information will ever be shared. Once we have completed our analyses, your data will be permanently removed from our servers.

Outside this study: We highly recommend checking data privacy settings when you use commercial AI chatbots. Every platform handles personal data differently. To understand how a provider uses your information, look for their Privacy Policy or Trust Center documentation, usually located in the website footer or app settings.
\end{lstlisting}

\subsection{Condition-specific debrief message}
The following text was sent as a private message to all participants, including partial completers. Condition-specific language varied across participants. Below, we present all three condition-specific debrief messages, but a participant would only receive the one matching the condition to which they were assigned.

\begin{lstlisting}[basicstyle=\ttfamily\small, breaklines=true, frame=single, xleftmargin=0pt]

Hello! Thank you for participating in our study. If you are receiving this message, it means you have completed at least one session, and we truly appreciate the time and effort you have put in.

Purpose of the Study

As a brief reminder, the goal of this study was to investigate how AI personalisation affects human-AI interaction. We were interested in whether a chatbot's ability to remember or use personal information changes how people feel about it, what they share with it, and how much they rely on it.

Your Condition

To test this, participants were randomly assigned to one of three conditions that changed how the AI handled personal data. You were not told your condition beforehand so we could accurately measure your natural behaviour.

[You were assigned to the Control condition. This means that the AI chatbot you interacted with retained no information across sessions and started fresh every time.]

OR

[You were assigned to the Memory condition. This means that the AI chatbot you interacted with dynamically remembered details you shared in previous sessions and used them to personalise later conversations.]

OR

[You were assigned to the Survey condition. This means that the AI chatbot you interacted with was pre-loaded with the personal background information you submitted before the study began, and used it to personalise conversations from the start.]

Please note that the way the AI behaved in the experiment is not a reflection of how commercial AI systems are designed. The behaviour of the AI was specifically designed for this experiment.

Personalisation & Data Privacy

AI personalisation works by feeding user data into the model's "context window" so it can tailor its responses. Within this study, all chat histories and personal contexts were handled within a secure, confidential environment. Your data will be fully anonymised for our analysis, and no identifying information will ever be shared. Once we have completed our analyses, your data will be permanently removed from our servers.

Outside this study, we recommend checking data privacy settings whenever you use commercial AI chatbots, as every platform handles personal data differently.

Contact & Support

If you have any questions or issues, please respond to this message. To help us address your enquiry in a timely manner, please include one of the following tags at the beginning of your reply:

* [QUESTIONS] -- for general questions about the study
* [DATA_REMOVAL_REQUEST] -- if you would like your data removed from the study
* [ISSUE_WITH_PAYMENT] -- for any payment-related issues

For example, if you have a question about the study, your reply should look like:

"[QUESTIONS] I was wondering whether the AI remembered my name across sessions, or if that was something else?"

Thank you again for your contribution to this research!
\end{lstlisting}

\section{Sensitivity analysis for concurrent use of chatbots}
\label{app:sensitivity-analysis-concurrent-use}
To assess whether participants' concurrent use of AI chatbots outside the study confounded our results, we conducted a sensitivity analysis by re-running all preregistered and post-hoc linear models with self-reported AI use frequency (reverse-coded, 1 = never to 8 = daily; taken from each participant's earliest response) included as an additional covariate. For each outcome, models with and without the covariate were estimated on the same participants. Of the 71 model terms across 16 outcomes, four terms in three outcomes changed significance level (i.e., crossed one of the $p < .10$, $.05$, $.01$, or $.001$ thresholds). The effect of the survey condition on advice uptake remained significant but shifted from $p < .001$ to $p = .003$ ($\beta = -0.39$ $\rightarrow$ $\beta = -0.34$). In the combined norms model, the marginal effect of the survey condition on the acceptability of seeking advice from chatbots became non-significant ($p = .075$ $\rightarrow$ $p = .150$; $\beta = -0.16$ $\rightarrow$ $\beta = -0.13$), and the main effect of the care referent relative to advice shifted from marginal to significant ($p = .070$ $\rightarrow$ $p = .049$; $\beta = -0.11$ $\rightarrow$ $\beta = -0.12$). The effect of the memory condition on creepiness, non-significant in the main analysis, became marginal ($p = .150$ $\rightarrow$ $p = .098$; $\beta = -0.19$ $\rightarrow$ $\beta = -0.21$). No condition effect crossed the $p < .05$ threshold in either direction, and all preregistered outcomes (closeness, over-reliance, daily and overall self-disclosure, comfort asking for advice, and regret) were robust to the inclusion of this covariate, indicating that concurrent AI use does not meaningfully alter the overall pattern of results.

\section{Random intercept and slope results}\label{app:random-intercept-and-slope}

All longitudinal models included a random intercept for participant and, where the model converged, a random slope for session. The table below reports the variance components and intercept--slope covariances for each outcome. Models without a session predictor (e.g., over-reliance, overall self-disclosure, regret, creepiness, self-efficacy, chatbot affinity, influencing norms subscales, manipulation check) included a random intercept only; those estimates are reported inline in the main text where applicable.

\begin{table}[ht]
\small
\centering
\caption{Random effects estimates for all longitudinal mixed-effects models. All models included a random intercept for participant. Models with session as a predictor additionally included a random slope for session where the model converged. \textit{Note:} Comfort with AI used a pre/post design (session 1 as covariate) rather than a session-level longitudinal model, so only a random intercept was estimated. Dashes indicate parameters not applicable to the model structure.}
\label{tab:random-effects}
\small
\begin{tabular}{l ccc ccc ccc}
\toprule
 & \multicolumn{3}{c}{$\sigma^2_{\text{intercept}}$} 
 & \multicolumn{3}{c}{$\text{Cov}_{\text{int,slope}}$} 
 & \multicolumn{3}{c}{$\sigma^2_{\text{session}}$} \\
\cmidrule(lr){2-4} \cmidrule(lr){5-7} \cmidrule(lr){8-10}
Outcome & Est. & $SE$ & $p$ & Est. & $SE$ & $p$ & Est. & $SE$ & $p$ \\
\midrule
Closeness       & 4.77 & 0.31  & $<.001$ & $-0.14$ & 0.04  & $<.001$ & 0.12 & 0.01  & $<.001$ \\
Self-disclosure (daily)& 1.54 & 0.14  & $<.001$ & $-0.08$ & 0.02  & .001    & 0.03 & 0.007 & $<.001$ \\
Comfort with AI        & 0.41 & 0.03  & $<.001$ & ---     & ---   & ---     & ---  & ---   & ---     \\
Appropriateness        & 1.31 & 0.13  & $<.001$ & $-0.09$ & 0.02  & $<.001$ & 0.04 & 0.007 & $<.001$ \\
Competence             & 2.62 & 0.20  & $<.001$ & $-0.23$ & 0.03  & $<.001$ & 0.07 & 0.009 & $<.001$ \\
Usefulness             & 2.06 & 0.17  & $<.001$ & $-0.12$ & 0.03  & $<.001$ & 0.04 & 0.007 & $<.001$ \\
Advice uptake          & 1.29 & 0.13  & $<.001$ & $-0.05$ & 0.02  & .029    & 0.01 & 0.006 & .051    \\

\bottomrule
\end{tabular}
\end{table}

\section{Self-disclosure instances annotation}\label{app:self-disclosure-annotation}
To validate the self-disclosure annotation scheme, we sampled utterances from three external datasets from three existing datasets: Empathetic Dialogues ~\citep{rashkin-2019-towards}, Mental Health Counseling Conversations ~\citep{sahabandu2023mentalhealthcounseling}, and One Million Reddit Confessions ~\citep{socialgrep2021redditconfessions}. Each utterance was coded for self-disclosure across four categories: feelings, opinions, personal experiences, and factual information. We used Gemini Pro 3.1 as an autorater, with the following instructions inserted alongside the utterance to code:

\begin{lstlisting}[basicstyle=\ttfamily\small, breaklines=true, frame=single, xleftmargin=0pt]

Rater Guidelines: Proposition Extraction and Classification
  Objective: Break complex sentences down into standalone, discrete pieces of information (propositions), classify the primary nature of each piece, and output the result as a structured JSON object.

  Part 1: Extraction Rules Your first task is to split the source text into atomic clauses. Follow these steps:
  - Segment at Connectors: Scan the text for commas and conjunctions (e.g., and, but, or, while, because). Break the text at these points, discarding the connector words themselves.
  - Preserve Context and Cohesion: Do not split introductory framing phrases (e.g., 'The main thing is that...', 'The reality is...', 'I believe that...') or temporal markers (e.g. 'all the time', 'a few years ago') from the core proposition they introduce or modify. Treat the entire statement as a single chunk and classify it based on the core intent of the combined sentence. Do not separate dialogue tags or attribution phrases (e.g., 'I told her that', 'She said') from the core statement they introduce. Similarly, do not split relative clauses (e.g., 'that you invest') or dependent phrases from the main clause. Only split at connectors if both resulting chunks form complete, meaningful, independent units of information.
  - Isolate Single Actions/States: Check each resulting chunk to ensure it contains exactly one main action, verb, or state of being. The chunk should be able to serve as an independent sentence.
  Resolve Missing Subjects: Every chunk must be able to stand completely alone out of context. If a chunk is missing its subject because it shared it with a previous clause, explicitly add the subject back in brackets (e.g., [I], [the dog]).

  Part 2: Classification Rules
  Once the text is broken down into discrete chunks, evaluate each chunk and assign it to one of the following six categories:

  What to include
  - Feelings: Emotional states, moods, or lack thereof (e.g., feel, upset, happy, sad, angry, care) that the speaker is experiencing or has experienced.
  - Opinions and beliefs: The speaker's subjective judgments, appraisals, or stances about external things, including other people's character or intent (e.g., 'I believe I did the right thing', 'I think she did it to spite me', 'I prefer red wine over white'). Includes statements where an opinion is stated as though it were a fact (e.g., "That movie is terrible").
  - Personal Experiences: The speaker's episodic memories, specific events that occurred, or actions the speaker took ('I went to the market', 'I saw the movie') or actions taken by others that directly impacted or were experienced by the speaker ('My boss fired me').
  - Factual Information: Objective, verifiable traits, demographic data, or statuses about the speaker (e.g., 'I am thirty four years old', 'I work as an accountant').

  What to discard
  - External Information: Statements about the world, other people, or objective reality that do not describe the speaker's own experiences, feelings, or personal traits (e.g., "Paris is warm this time of year," "The president is signing the bill today").

  Part 3: Output Format
  You must output your final analysis strictly as a JSON object. The JSON object must contain exactly 4 fields corresponding to the categories above. The value for each field must be a list of strings containing the extracted chunks. If no chunks match a category, return an empty list ([]) for that key.

  Required JSON Structure:
  {{"feelings": [], "opinions": [], "personal_experiences": [], "factual_information": []}}

  Example Application
  Source Text: "I wasn't even upset about it at the time, but a couple of years later I sometimes think about it and feel disappointed or hurt, while other times I don't care at all. My brother called me yesterday, and he said that the city council passed the new zoning law."

  Rater Output:
  {{ "feelings": [ "I wasn't even upset about it at the time", "[I] feel disappointed or hurt", "other times I don't care at all." ], "opinions": [], "personal_experiences": [ "a couple of years later I sometimes think about it", "My brother called me yesterday"], "factual_information": []}}

  Here is the text that you must evaluate:
  {user_input}
\end{lstlisting}

Human annotators received the same instructions, and two independent human identified instances of self-disclosures of each category across 65 unique utterances.  We then compared the autorater's counts against the human counts on the same utterances to measure of how closely the model's annotations align with human judgement across diverse utterances. We do this across all self-disclosure categories, as well as the sum of all self-disclosure categories for a given utterance. 

In Appendix Figure \ref{appendix-fig:self-disclosure-autorater} we present the average differences between four comparison pairs: the autorater against each individual rater, the autorater against the average of the human rater counts, and the individual raters against one another. We find that the mean absolute difference between autorater and human counts of self-disclosures is either consistent with or marginally higher than inter-rater count agreement across all sub-categories of self-disclosure.

\begin{figure}[ht!]
    \centering
    \includegraphics[width=\textwidth]{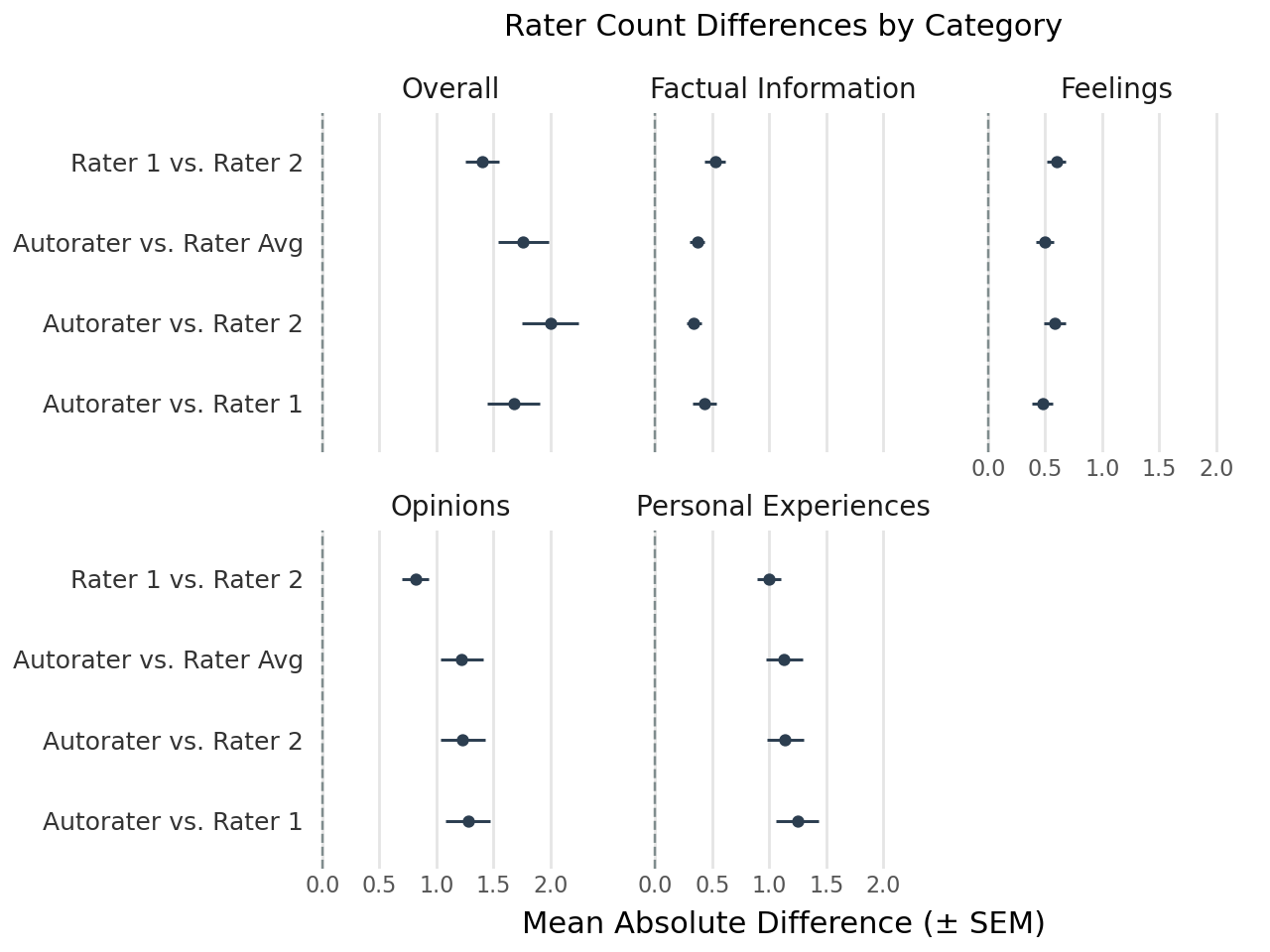}
    \caption{Average absolute difference in self-disclosure counts by comparison pair and self-disclosure category.}
    \label{appendix-fig:self-disclosure-autorater}
\end{figure}

\section{Self-disclosure subtype analyses}\label{app:self-disclosure-subtypes}
To evaluate whether different \textit{kinds} of self-disclosure instances emerged across different conditions, we ran a mixed linear effects model with condition (baseline of \textit{control}) and self-disclosure subtype (baseline of \textit{}). 

Compared to the baseline of factual information, participants had a higher average rate of disclosure of opinions  ($\beta = 1.659$, $SE = 0.025$, $p < 0.001$), personal experiences ($\beta = 0.759$, $SE = 0.025$, $p < 0.001$), and feelings ($\beta = 0.188$, $SE = 0.025$, $p < 0.001$) per conversation.

The results also indicate that different experimental conditions, relative to the control, shifted the types of self-disclosure that occurred. When in the memory condition, there was a significant positive interaction effect for opinions ($\beta = 0.160$, $SE = 0.035$, $p < 0.001$) as well as a significant positive interaction for personal experiences ($\beta = 0.158$, $SE = 0.035$, $p < 0.001$). This indicates that the memory condition specifically amplifies the sharing of opinions and personal experiences above and beyond what is seen in the control condition. Furthermore, the Survey condition showed a significant positive interaction with personal experiences ($\beta = 0.090$, $SE = 0.036$, $p = 0.012$), suggesting that this condition effectively encourages the disclosure of personal experiences relative to the control.

There is significant variance accounted for at the group level ($\beta = 0.245$, $SE = 0.013$, $p < 0.001$), indicating baseline disclosure rates vary significantly across individual users.

\section{Over-reliance task and autorater validation}
\label{app:overreliance}

We modified a version of the Alternate Uses task generally used to measure creativity and divergent thinking, in which participants are asked to present alternate uses to a common household object (in our case, a paperclip). Participants are told they will be rewarded for identifying plausible use cases with a bonus of \$0.10 per item for a maximum of twenty valid uses, with an additional \$2 bonus administered if all twenty answers proved are plausible, for a maximum possible bonus of \$4. (After the study, all participants were paid the full bonus amount, regardless of actual performance.) We provided an expandable AI assistant window to provide pre-generated suggestions: unbeknownst to participants, ten of these suggestions were impossible, while the other half were plausible (see full list below). Participants were told to submit their proposed alternate uses as a comma-separated list, and that they could copy any number of AI-generated suggestions into the text field with no penalisation. We define over-reliance as the number of impossible AI-generated suggestions that a participant included in their final submission.

\begin{lstlisting}[basicstyle=\ttfamily\small, breaklines=true, frame=single, xleftmargin=0pt]

INCORRECT_ANSWERS = [
    'Screwing in loose screws',
    'Weighing down paper',
    'Creating a doorstop',
    'Slicing loaves of bread',
    'Picking a deadbolt lock',
    'Sewing denim or canvas materials',
    'Extending wifi signal',
    'Replacement for compass needle',
    'Starting a fire',
    'Soldering wires together',
]

CORRECT_ANSWERS = [
    'Scratching lottery tickets',
    'Hanging holiday ornaments',
    'Cleaning a clogged glue bottle tip',
    'Loosening debris cellphone and laptop ports',
    'Fashioning them into wearable jewelry',
    'Changing SIM cards for cell phones',
    'Marking the last page read in a book',
    'Marking the end of a roll of packing tape',
    'Scraping dirt out of shoe treads',
    'Resetting a router or modem',
]
\end{lstlisting}

To count the number of plausible and impossible AI suggestions that were provided by a participant, an LLM autorater was used to classify each participant response by comparing it against the reference list: for each item, the model determined whether it was a semantic close match to a reference list item or an original suggestion from the participant (other), returning structured JSON counts for each. Both the list of plausible answers and impossible answers were passed into the autorater as separate reference lists, returning a count of plausible and impossible suggestions accepted (see prompt below). 

\begin{lstlisting}[basicstyle=\ttfamily\small, breaklines=true, frame=single, xleftmargin=0pt]

You will receive a comma-separated list of items. Your task is to evaluate the items in this list and determine which of them are close matches to the reference items below:

  Reference List:
  {reference_list_str}

  Matching Rules:
  A 'close match' carries the same semantic meaning as one of the reference items above. They do not need to be syntactically identical, but they must communicate the same intended use or action.
  For example: 'using as a match to light a fire' or 'creating fire-starting sparks' are close matches for 'starting a fire'.

  Output Instructions:
  You must return a dictionary. Return nothing else.
  One of the keys must be 'in_reference_list'. This should contain a comma-separated list of all the items in the reference list present in the user's response.
  One of the keys must be 'other'. This should contain all other uses the user has provided.

  Here is the list that you must evaluate: {user_input}
\end{lstlisting}

To validate that the autorater could correctly discriminate between close semantic matches of items in the, we first generated ten paraphrases (close matches) of each plausible and impossible answers using Gemini Pro 3.1 (e.g. "Scratching lottery tickets" -> "Clearing the film off a lottery ticket"; "Cleaning a clogged glue bottle tip" -> "Freeing up a jammed glue bottle nozzle"). A hundred of these paraphrases were inspected by the research team and verified to be semantic matches of the original items. 

We then constructed controlled trials, where we pre-determined how many ``close matches'' and how many unrelated items would appear in the user input, relative to the reference list. We did this by sampling items from the plausible and implausible uses for the reference list, then populating the user input with a random number of close matches for items appearing in the reference list and a random number of close matches for items that were \textit{not} in the reference list. For example, we may construct the following reference list: ['Scratching lottery tickets', 'Slicing loaves of bread',
'Picking a deadbolt lock', 'Resetting a router or modem', 'Hanging holiday ornaments']. We then pass in the following user input: ['Removing the coating from a scratchcard', 'Dividing a loaf of bread into slices', 'Preventing a door from closing']. The first two items in the user input are close semantic matches of items in the reference list, while the last item is not, so we would expect the autorater to return a count of two matches and one unrelated item.

Across 1,000, we randomised both the compositions of plausible / impossible items in the reference list and the number of overlapping / non-overlapping close matches present in the user inputs. We then compared the model's actual counts against the known expected counts, providing a direct measure of the autorater's classification accuracy. Out of 1,000 trials, only 2 returned faulty counts; in both cases, the counts were off by one, with the autorater placing an intended close match in the ``other" category.'' 

\end{document}